\documentclass[10pt,journal]{IEEEtranTCOM}

\usepackage[english]{babel}
\usepackage[usenames]{color}
\usepackage[cp1250]{inputenc}
\usepackage{amsfonts}
\usepackage{amsthm}
\usepackage{graphicx}
\usepackage{xcolor}
\usepackage{epsfig}
\usepackage{mathrsfs}
\usepackage{amsmath}
\usepackage{algorithm}
\usepackage{algorithmic}
\usepackage{hyperref}
\usepackage{stackrel,amssymb}

\theoremstyle{plain}

\begin{document}
\newcommand{\bea}{\begin{eqnarray}}
\newcommand{\eea}{\end{eqnarray}}
\newcommand{\be}{\begin{equation}}
\newcommand{\ee}{\end{equation}}
\newcommand{\beas}{\begin{eqnarray*}}
\newcommand{\eeas}{\end{eqnarray*}}
\newcommand{\bs}{\backslash}
\newcommand{\bc}{\begin{center}}
\newcommand{\ec}{\end{center}}
\def\SC {\mathscr{C}}
\def\ind {\perp\!\!\!\perp}

\title{Biology-inspired joint distribution neurons \\ based on Hierarchical Correlation Reconstruction \\ for multidirectional propagation of values and densities }
\author{\IEEEauthorblockN{Jarek Duda}, 
\IEEEauthorblockA{Jagiellonian University, Cracow, Poland, \emph{jaroslaw.duda@uj.edu.pl}}}
\maketitle

\begin{abstract}
Recently a million of biological neurons (BNN) has turned out better from modern RL methods in playing Pong~\cite{RL}, reminding they are still qualitatively superior e.g. in learning, flexibility and robustness - suggesting to try to improve current artificial e.g. MLP/KAN for better agreement with biological. There is proposed extension of KAN approach to neurons containing model of local joint distribution: $\rho(\mathbf{x})=\sum_{\mathbf{j}\in B} a_\mathbf{j} f_\mathbf{j}(\mathbf{x})$ for $\mathbf{x} \in [0,1]^d$, adding interpretation and information flow control to KAN, and allowing to gradually add missing 3 basic properties of biological: 1) biological axons propagate in both directions~\cite{axon}, while current artificial are focused on unidirectional propagation - joint distribution neurons can repair by substituting some variables to get conditional values/distributions for the remaining. 2) Animals show risk avoidance~\cite{risk} requiring to process variance, and generally real world rather needs probabilistic models - the proposed can predict and propagate also distributions as vectors of moments: (expected value, variance) or higher. 3) biological neurons require local training, and beside backpropagation, the proposed allows many additional ways, like direct training, through tensor decomposition, or finally local and promising: information bottleneck. Proposed approach is very general, can be also used as extension of softmax in embeddings of e.g. transformer, JEPA, Mamba, suggesting interpretation that features are mixed moments of joint density of real-world properties.
\end{abstract}
\textbf{Keywords}: artificial/biological neural networks, machine learning, Kolmogorov-Arnold Network, joint/conditional distribution, tensor decomposition, mutual information, information bottleneck, HSIC, softmax, embedding, transformer, JEPA, Mamba
\section{Introduction}
Just connecting a million of biological neurons to electrodes, turned out better from modern Reinforcement Learning method in playing Pong~\cite{RL} or recently Doom~\footnote{https://www.newscientist.com/article/2517389-human-brain-cells-on-a-chip-learned-to-play-doom-in-a-week/}, reminding that biological neurons still have qualitative superiority, especially in learning, flexibility and robustness, summarized further in Fig. \ref{table}. There are low level approaches to catch up like spiking neural networks~\cite{spiking}, however, with energy efficiency as nearly the only success. Biological neurons might be too complicated for practical direct modelling: they work in bursts of spikes, which annihilate when meet, traveling through complex networks in all directions.

\begin{figure}[t!]
    \centering
        \includegraphics[width=9cm]{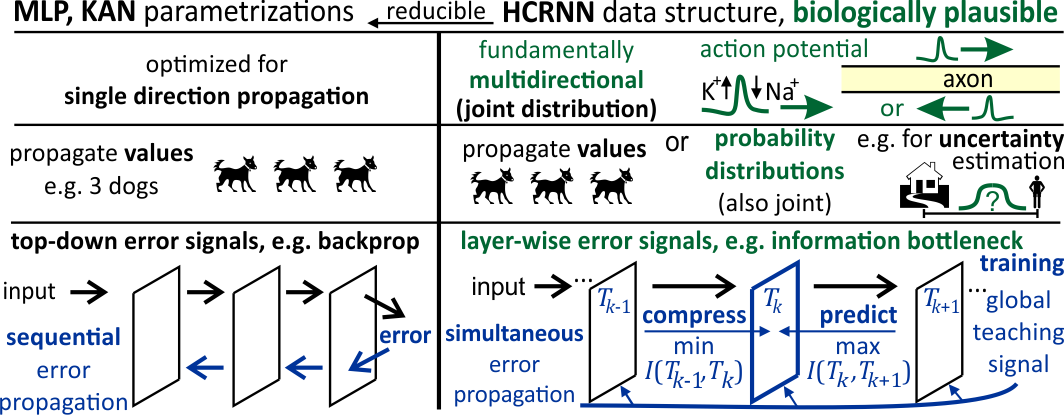}
        \caption{Like in MLP/KAN approach, we search for \textbf{logical neurons} supposed to extract essential mathematics hidden in biological neurons - focusing on their usually missing 3 properties: bidirectional propagation~\cite{axon}, also working of distributions e.g. for observed risk avoidance~\cite{risk}, and using local training like information bottleneck~\cite{information}. All 3 are natural for \textbf{neurons containing joint distribution model}, practical with HCR approximating joint density as a linear combination: $\rho(\mathbf{x})=\sum_{\mathbf{j}\in B} a_\mathbf{j} f_\mathbf{j}(\mathbf{x})$ for $a_\mathbf{j}$ moments as neuron parameters. }
        \label{learn}
\end{figure}
\begin{figure}[t!]
    \centering
        \includegraphics[width=9cm]{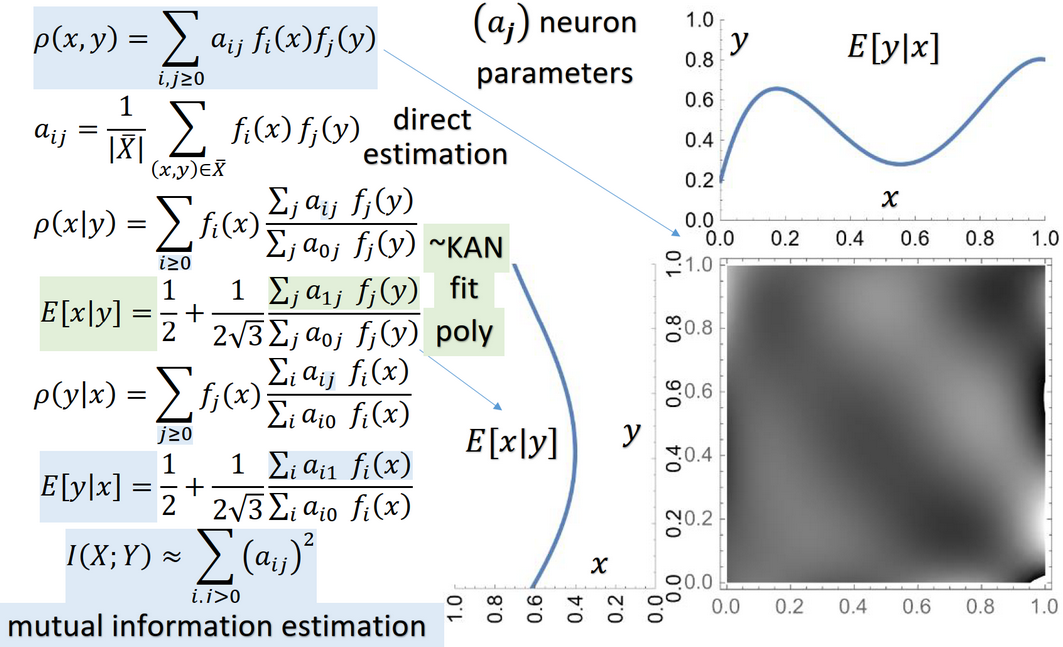}
        \caption{Basic formulas and example for $d=2$ variables HCR neuron, using convenient variable normalization to nearly uniform in $[0,1]$. \textbf{Neuron contains matrix of moments:} $a_{ij}$ (generally order $d$ tensor), allowing to propagate in various directions by substituting some variables and normalizing to get estimated conditional density for the remaining - just permuting indexes to change propagation direction. For value propagation we can take expected value, what restricts prediction to 1-st moment - becoming summation of trained polynomials like in KAN\cite{kan}, allowing to view HCRNN as extension of KAN e.g. for optional density propagation and change of propagation direction. 
        Additionally, such $a_{ij}$ parameters are actual moments allowing for better interpretation, and inexpensive estimation of mutual information - for training or monitoring of information flow.}
        \label{2d}
\end{figure}
Therefore, to catch up with BNNs, it seems more promising to search for higher level MLP\cite{mlp}/KAN\cite{kan}-like  \textbf{logical neurons} - trying to extract mathematics hidden in biological neurons, asking if it is e.g. matrix multiplication-like. To understand this mathematics, we focus on 3 fundamental differences, summarized in Fig. \ref{learn}: uni-directional vs bi-directional propagation of biological axons~\cite{axon}, working only on values vs also on distributions - observed in animals e.g. as risk avoidance ~\cite{risk}, and finally BNNs need local training approaches like looking the most promising: information bottleneck~(\cite{information,information1, information2}).

\textbf{Neurons containing local joint distribution model}  would allow to include above features of BNNs, by substituting some variables allows to predict and propagate conditional probabilities - what seems crucial for animals to operate in complex physical world, in practice being probabilistic. If only in reach at a reasonable cost, evolution should provide organisms ways to predict conditional distributions, preferably also propagating uncertain information through neural networks.

\begin{figure}[t!]
    \centering
        \includegraphics[width=9cm]{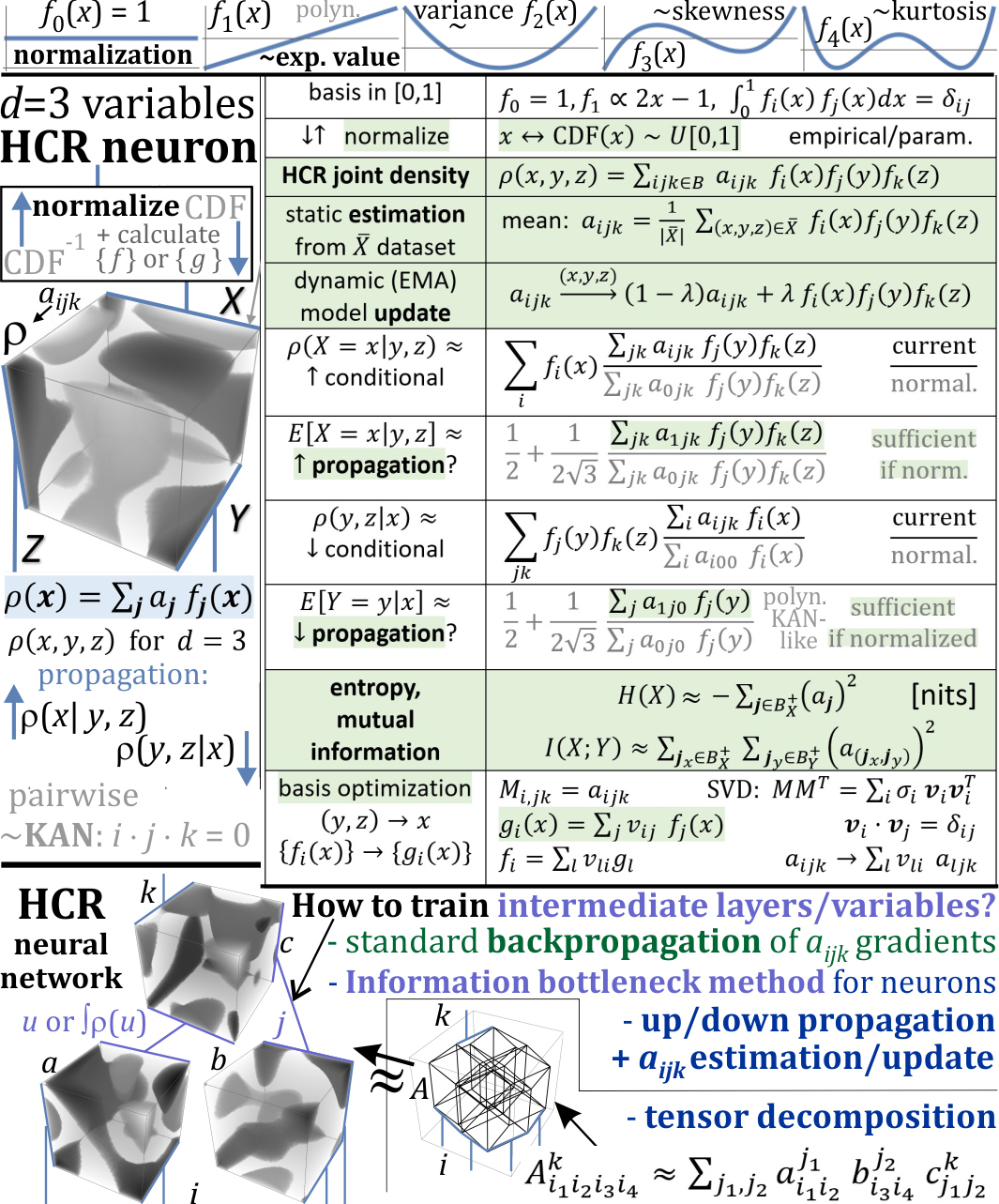}
        \caption{The proposed HCR neuron and neural network (HCRN, HCRNN) containing local joint distribution model represented in $(a_\mathbf{j})_{\mathbf{j}\in B}$ tensor, e.g. $(a_{ijk})_{i,j,k\in \{0,..,m\}}$ for $d=3$ connections. \textbf{Top}: used orthonormal polynomial basis for uniform weight in $[0,1]$, convenient for such normalization to quantiles.\\ 
        \textbf{Middle}: HCR neuron containing and applying joint distribution model like in Fig. \ref{2d}, but for $d=3$ variables, and gathered formulas for direct estimation/model update, entropy/mutual information, conditional densities and their expected values for propagation. Such $\rho$ density parametrization can drop below 0, what is usually repaired by calibration e.g. using normalized $\max(\rho,0.1)$ density. However, for neural networks its approximations should be corrected by training, hence we neglect calibration essentially simplifying calculations to shown formulas. Propagating only expected values and normalizing, we can use only the marked nominators - as in KAN optimizing nonlinear functions (polynomial here) by including only pairwise dependencies ($a$ with two nonzero indexes), extending to their products to consciously include higher order dependencies. \\ \textbf{Bottom}: schematic HCR neural network and some training approaches of intermediate layers - which in HCR can be treated as values or their distributions (replacing $f_i(u)$ with its $i$-th moment: $\int_0^1 \rho(u) f_i(u) du$). There is also visualized tensor decomposition approach - estimate dependencies (e.g. pairwise) for multiple variables and try to automatically decompose it to multiple dependencies of a smaller numbers of variables with algebraic methods.    
        }
        \label{intr}
\end{figure}

However, modelling joint distributions is generally quite difficult - for practicality needs to be simplified as possible, like representation of \textbf{joint density as just a linear combination} like polynomial, of conveniently normalized variables - we refer to as HCR, leading to simple formulas with e.g. just matrix/tensor products as in Fig. \ref{2d}, \ref{intr} - relatively inexpensive calculations, suggesting biological neurons might hide similar representation. 

\begin{figure}[t!]
    \centering
        \includegraphics[width=9cm]{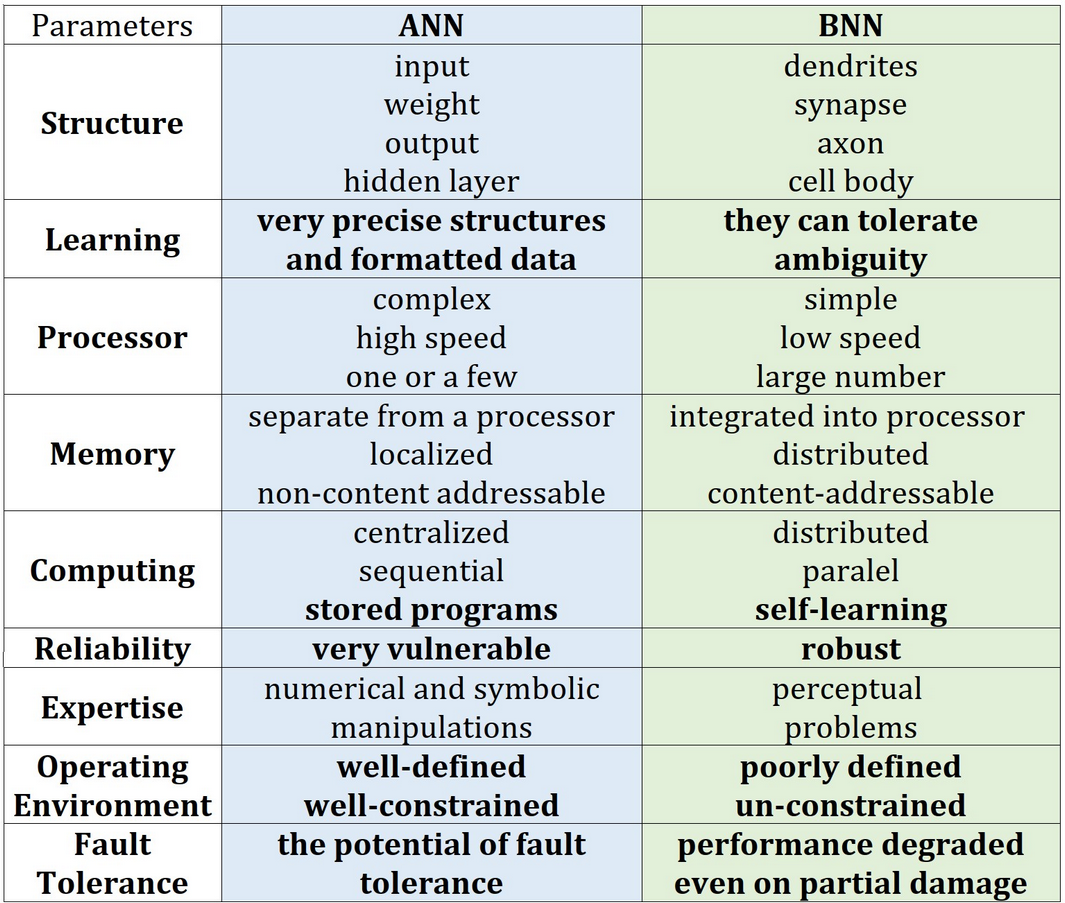}
        \caption{Summary of differences between artificial (ANN) and biological neural networks (BNN, based on \url{https://www.geeksforgeeks.org/difference-between-ann-and-bnn/}) - \textbf{BNNs are qualitatively superior in terms of learning, flexibility and robustness - just increasing the number of neurons might be insufficient to catch up}. To build ANNs closer to BNN capabilities, we should include their neuron-level properties, summarized in Fig. \ref{learn}, all \textbf{possible for neurons containing local joint distribution model} - allowing to build models of the world, which in practice is probabilistic.}
        \label{table}
\end{figure}
\begin{figure}[t!]
    \centering
        \includegraphics[width=9cm]{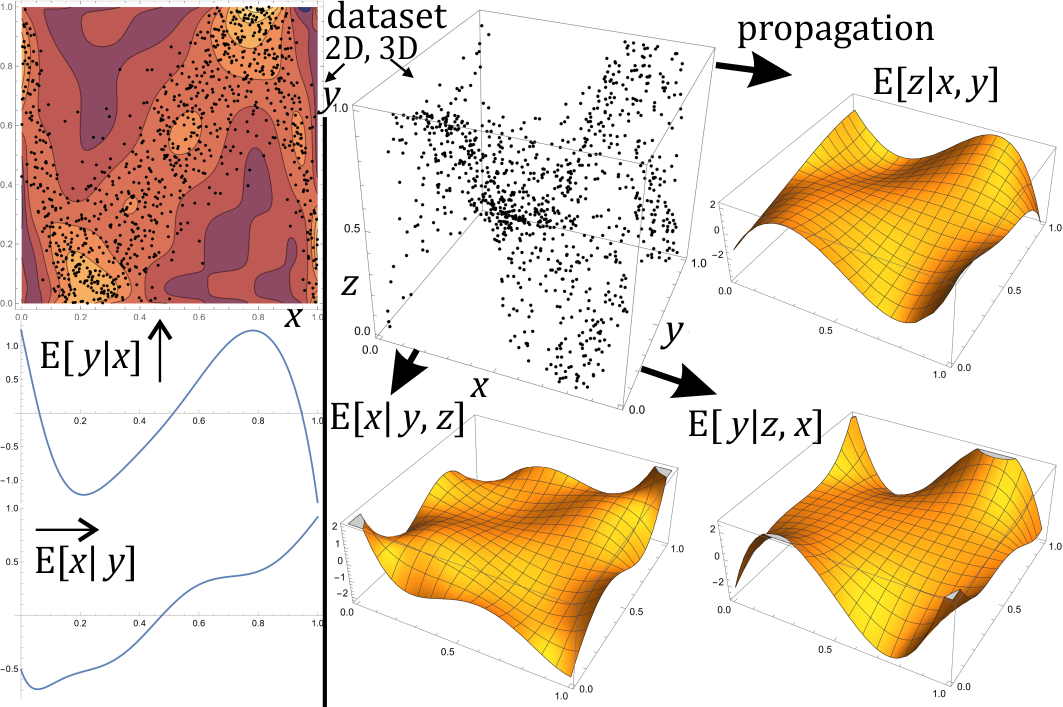}
        \caption{Simple 2/3D examples from \href{https://community.wolfram.com/groups/-/m/t/3241700}{HCRNN Wolfram notebook} of propagation in any direction based on the shown datasets (points) as conditional expected values, here being degree $m=8$ polynomials. }
        \label{HCRNN}
\end{figure}

HCR stands for Hierarchical Correlation Reconstruction, and has already found many applications~(\cite{hcr0,hcr1,hcr2,hcr3,hcr4,hcr5}) - allowing to hierarchically decompose dependencies into mixed moments $a_\textbf{j}$, predict and propagate them - here extended into neural networks as HCRNN \footnote{HCR introduction: https://community.wolfram.com/groups/-/m/t/3017754, HCRNN application: https://community.wolfram.com/groups/-/m/t/3241700} we focus on in this article.

While this approach can be also viewed as just parametrization, also allowing backpropagation training, it additionally allows many other training ways - briefly summarized in the bottom of Fig. \ref{intr}. E.g. having all neuron connections we could directly estimate $a_\textbf{j}$ moments between them, or update their current model. Working on joint distributions with easy to estimate mutual information, we can use mentioned local information bottleneck approach - offering many improvements over previously used HSIC~\cite{hsic1,hsic2}, like much lower cost and better sensitivity. There is also possible training by tensor decomposition, extending SVD (singular value decomposition) - trying to automatically decompose larger joint distribution into multiple smaller ones.

The discussed approach is very general - allowing e.g. to predict or propagate densities as approximated vectors of moments: (expected value, variance) and optionally higher like skewness, kurtosis. It can be also used to modify standard approaches e.g. from value to density prediction, or softmax embedding-based like transformers: usually they predict single energy-like coefficient $E$ to define probability with softmax: $\textrm{Pr}\propto \exp(-E)$. However, if in embedding it represents e.g. age, single feature seems insufficient to describe our incomplete knowledge - in contrast to extension to features being (mixed) moment-like coefficients $a_\mathbf{j}\sim f_\mathbf{j}(\mathbf{x})$ describing (joint) densities of real properties.

This article introduces to HCR from perspective of neural network applications, earlier suggested in \cite{hcr0}, to be extended in future e.g. with practical realizations replacing MLP, KAN, or to extend current embedding-based architectures, like transformers, JEPA or Mamba to work with density-based (un)embedding.

\section{HCR neural networks (HCRNN)}
This main Section introduces to HCR and discusses it as a basic building block for nextgen neural networks.

\begin{figure}[t!]
    \centering
        \includegraphics[width=9cm]{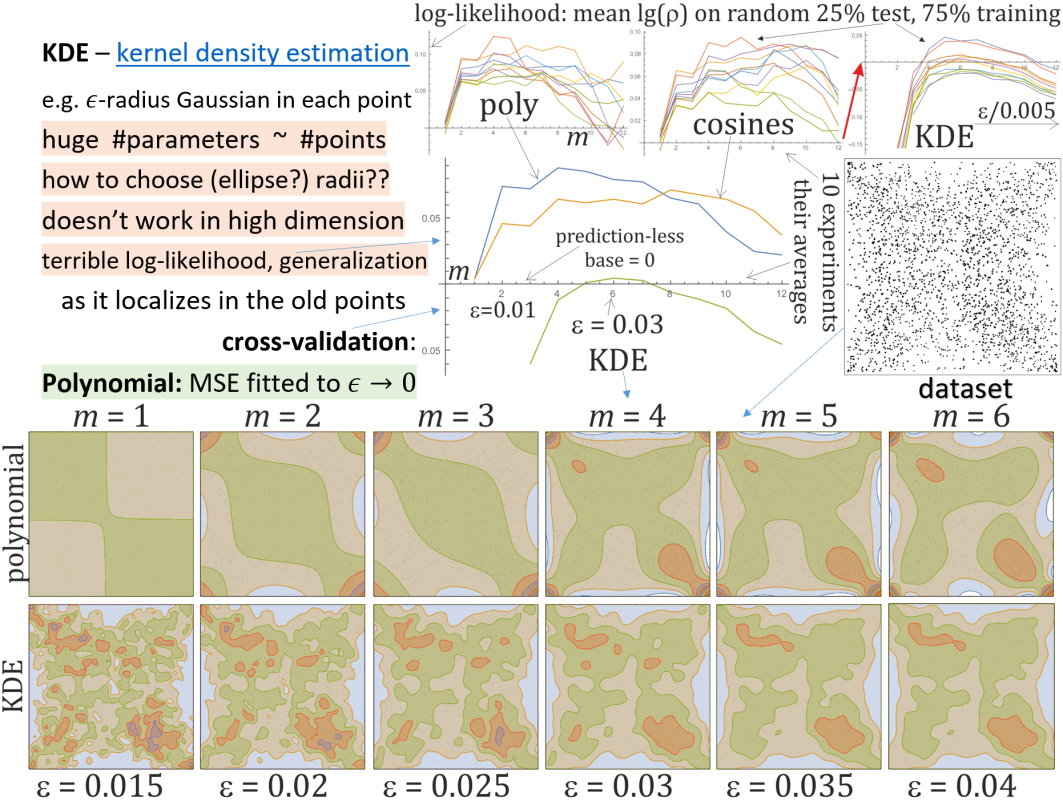}
        \caption{2D example comparison of \textbf{local basis KDE} (kernel density estimation) vs \textbf{global basis HCR} (available code in \href{https://community.wolfram.com/groups/-/m/t/3017754}{HCR Wolfram notebook}) modelling joint density for  dataset as points shown on the right. Assuming  $\rho=1$ trivial joint density, we would get 0 log-likelihood evaluation (mean $\lg(\rho(x))$ over dataset). Training on a randomly chosen subset and calculating log-likelihood on the remaining subset (cross-validation), we can see local KDE barely exceed this 0 for trivial model (best for $\epsilon\approx 0.03$ kernel width), while global: using polynomial or trigonometric basis easily exceed it - they are able to extract crucial generalizing features. The highest found log-likelihood was for $m=4$ polynomials - using popular 4 moments: expected value, variance, skewness and kurtosis. Intuitively, KDE assumes that new points will be close to the old points - what poorly generalizes, in contrast to global features like moments used by HCR.  }
        \label{local}
\end{figure}

\begin{figure}[t!]
    \centering
        \includegraphics[width=9cm]{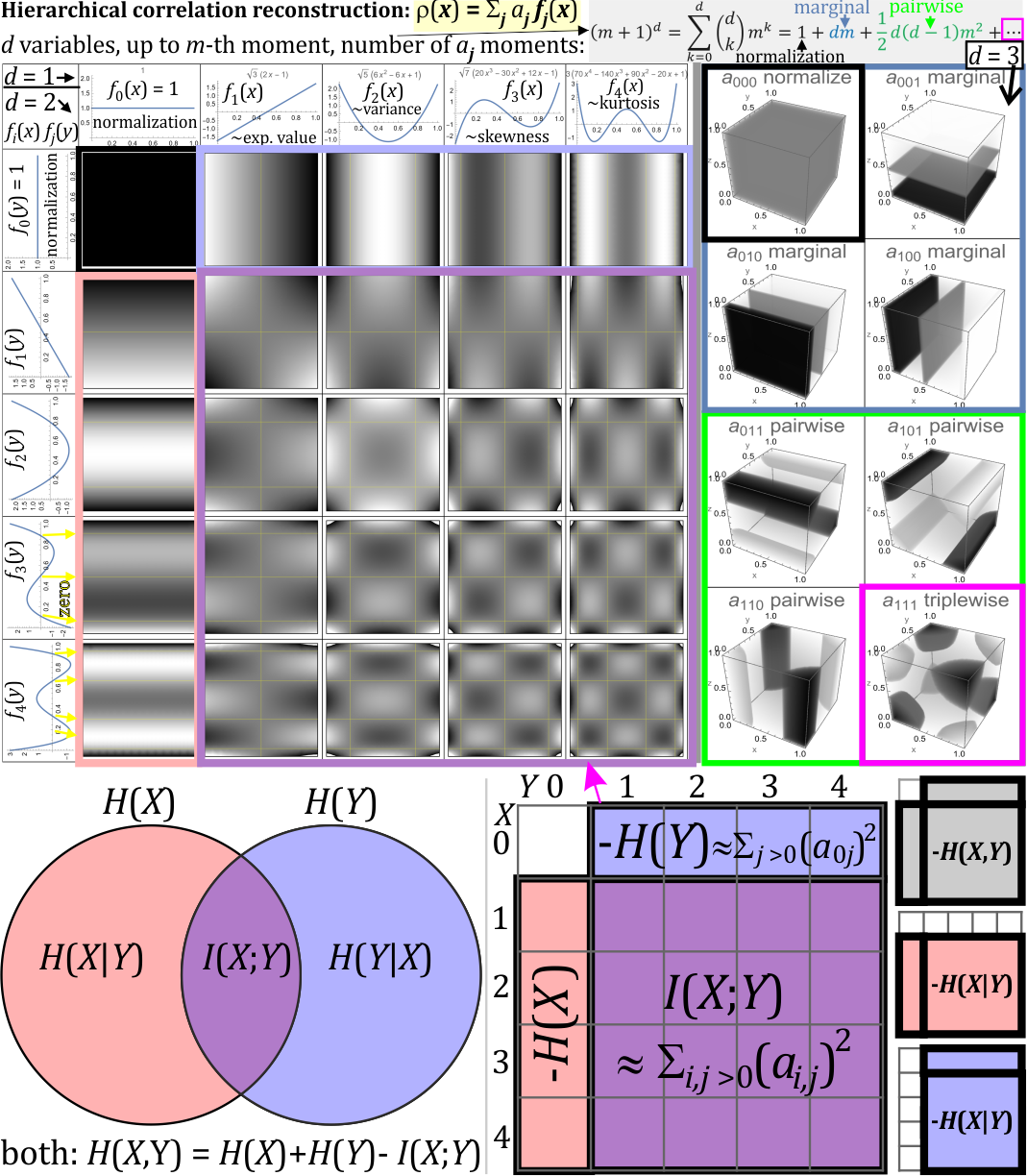}
        \caption{\textbf{Top}: Visualized part of HCR polynomial $[0,1]$ basis in $d=1$ dimension and $f_\textbf{j}(\textbf{x})=\prod_{i=1}^d f_{j_i}(x_i)$ product bases for $d=2, 3$. E.g. for $d=3$ the assumed joint density becomes $\rho(x,y,z)=\sum_{ijk} a_{ijk} f_i(x) f_j(y) f_k(z)$. As $f_0=1$, zero index in $a_{ijk}$ means independence from given variable, hence $a_{000}=1$ corresponds to normalization, $a_{i00},a_{0i0},a_{00i}$ for $i\geq 1$ describe marginal distributions through $i$-th moments. Then $a_{ij0},a_{i0j},a_{0ij}$ for $i,j\geq 1$ describe pairwise dependencies through mixed moments, and finally $a_{ijk}$ for $i,j,k\geq 1$ describe triplewise dependencies. This way we literally get \textbf{hierarchical correlation reconstruction through moments describing dependencies between increasing numbers of variables}, with clear interpretation of coefficients of e.g. trained HCR-based neural network. 
        \textbf{Bottom}: discussed further information theoretic view, allowing inexpensive estimation of e.g. mutual information as sum of square of nontrivial moments between two variables. }
        \label{HCR}
\end{figure}

\subsection{Introduction to Hierarchical Correlation Reconstrution}
As in copula theory~\cite{copula}, it is convenient to use \textbf{normalization} of variables to nearly uniform distribution in $[0,1]$. It requires transformation through cumulative distribution function (CDF): value $x\to \textrm{CDF}(x)$ becomes its estimated quantile, e.g. $1/2$ for median. This CDF can be modeled with some parametric distribution using parameters estimated from dataset e.g. $x\to$ $$\textrm{CDF}_{N(0,1)}((x-E[X])/\sqrt{\textrm{var}(X)})=
\textrm{CDF}_{N\left(E[X],\sqrt{\textrm{var}(X)}\right)}(x)$$ for Gaussian distribution, or can be empirical distribution function (EDF): $x$ becomes its position in dataset rescaled to $(0,1)$. For neural networks such normalization is usually made in batches~\cite{batch}, here needed to be used between layers (can be skipped in further linearization). In practice should be nearly constant between batches, approximated, parameterized e.g. by Gaussian, put into tables, inversed for backward propagation, etc.\\

For $d$ normalized variables: $\mathbf{x}\in [0,1]^d$, in HCR we \textbf{represent joint distribution as a linear combination}, conveniently in some product basis $B=B^+\cup \{\textbf{0}\}$ with $f_0(x)=f_\mathbf{0}(\mathbf{x})=1$:
\be \rho(\mathbf{x})=1+\sum_{\mathbf{j}\in B^+} a_{\mathbf{j}}f_{\mathbf{j}}(\mathbf{x})=\sum_{\mathbf{j}\in B} a_{\mathbf{j}}f_{\mathbf{j}}(\mathbf{x})=\sum_{\mathbf{j}\in B} a_{\mathbf{j}} \prod_{i=1}^d f_{j_i} (x_i)\label{model}\ee
where $B^+=B\backslash \{\textbf{0}\}$ removes zero corresponding to normalization as $f_0=1$, bold fonts denote vectors: $\mathbf{j}=(j_1,..,j_d)$.

Assuming orthonormal basis: $\int_0^1 f_i(x) f_j(x) dx=\delta_{ij}$ \textbf{static estimation}~\cite{rapid} (minimizing mean-squared error between kernel density estimation smoothed sample and discussed parametrization) from $\bar{X}$ dataset becomes just:
\be a_{\mathbf{j}} =\frac{1}{|\bar{X}|} \sum_{\mathbf{x}\in \bar{X}} f_{\mathbf{j}}(\mathbf{x})=  \frac{1}{|\bar{X}|} \sum_{\mathbf{x}\in \bar{X}} \prod_{i=1}^d f_{j_i} (x_i) \label{estim}\ee 
We assume here \textbf{orthonormal polynomial basis} (rescaled Legendre), allowing to interpret coefficients as moments of normalized variables, being approximately cumulants, starting with:
\be f_0 = 1\qquad\textrm{corresponds to normalization}\label{leg}\ee
$$ f_1(x)=\sqrt{3} (2 x-1)\qquad \sim\textrm{expected value}$$
$$f_2(x) =\sqrt{5} \left(6 x^2-6 x+1\right) \qquad \sim\textrm{variance} $$
$$ f_3(x) = \sqrt{7} \left(20 x^3-30 x^2+12 x-1\right)\qquad\sim\textrm{skewness}$$
$$ f_4(x) =3 \left(70 x^4-140 x^3+90 x^2-20 x+1\right)\qquad\sim\textrm{kurtosis} $$

Independent $a_\mathbf{j}$ estimation as just average allows to control uncertainty of the found parameters $\propto 1/\sqrt{|\bar{X}|}$.

Alternatively we could use various trigonometric bases (e.g. discrete cosine/sine transform DCT/DST) - especially for periodic variables, localized like B-splines used by the original KAN, from wavelets, finite elements methods. Instead of normalization to uniform in $[0,1]$, we could use a different one e.g. to Gaussian distribution, times Hermite polynomials for orthonormal basis. For discrete variables we can use one-hot encoding, or e.g. its SVD/CCA-based optimization as in \cite{hcr5}.

As in Fig. \ref{HCR} and $f_0=1$, $a_{\mathbf{j}}$ coefficients are mixed moments of $\{i:j_i\geq 1\}$ variables of nonzero indexes, independent from variables of zero indexes, allowing for literally \textbf{hierarchical decomposition} of statistical dependencies: start with  $a_{0..0}=1$ for normalization, add single nonzero index coefficients to describe marginal distributions, then add pairwise dependencies with two nonzero indexes, then triplewise, and so on. For example $a_{2010}$ coefficient would describe dependence between 2nd moment of 1st variable and 1st moment of 3rd variable among $d=4$ variables. Generally the selection of basis $B$ is a difficult question, e.g. to use only pairwise dependencies up to a fixed moment $m$, preferably optimized during training, maybe separately for each neuron or layer. Such decomposition also allows to efficiently work with \textbf{missing data} by using to estimate/update/propagate only $a_\mathbf{j}$ coefficients with zero indexes for the missing variables, as $j_i=0$ zero index means independence from given variable.

While static estimation averages over dataset with equal weights, for \textbf{dynamic updating} we should increase weights of recent values, e.g. using computationally convenient exponential moving average (EMA) for some memory parameter $0\ll\eta < 1$, which can be extended to trained coefficient dependent  $\eta_{\mathbf{j}}$, or even matrices as in Mamba~\cite{mamba} discussed in Section \ref{mamb}:
\be a_{\mathbf{j}} \xrightarrow{\mathbf{x}=(x_1,..,x_d)}  \eta a_{\mathbf{j}} +(1-\eta)  \prod_i f_{j_i} (x_i)\label{update}  \ee

However, modelling (joint) probability density as a \textbf{linear combination can sometimes lead to negative densities} - to avoid this issue, there is usually used \textbf{calibration}: instead of the modelled density $\rho$, use e.g. $\varphi(\rho)=\max(\rho,0.1)$ or softplus $\varphi(\rho)=\alpha\ln(1+\beta\exp(\gamma\rho))$ for some $\alpha,\beta,\gamma>0$, and divide by integral to remain normalized density. However, it makes computations much more complex and costly, especially normalization, in higher dimensions - for neural network applications we usually should be able to ignore this issue to simplify calculations, especially working on expected values and normalizing between layers - its approximations should be compensated in training. Therefore, we mostly neglect this issue/calibration in this article, however, it should be remembered and is discussed in Section \ref{calsec}, maybe including calibration with proportional normalization over possibilities like softmax:  \be\textrm{Pr}(\mathbf{x})= \frac{\varphi(\rho(\mathbf{x}))}{\sum_{\mathbf{y}\in X} \varphi(\rho(\mathbf{y}))}\propto \varphi(\rho(\mathbf{x}))\label{propo}\ee
\subsection{Conditional distributions and expected value propagation}
Having (\ref{model}) model of joint distribution, to get \textbf{conditional distribution} we need to substitute known variables, and normalize dividing by integral, being 0-th coefficient to make $a_0=1$ : 
$$ \rho(x_1|x_2,\ldots,x_d) =
\frac{\sum_{\mathbf{j}} a_{\mathbf{j}} f_{j_1}(x_1)f_{j_2}(x_2)\ldots f_{j_d}(x_d)}
{\int_0^1 \sum_{\mathbf{j}} a_{\mathbf{j}} f_{j_1}(x_1)f_{j_2}(x_2)\ldots f_{j_d}(x_d) dx_1}=
$$
\be=\sum_{j_1} f_{j_1}(x_1) 
 \frac{\sum_{j_2\ldots j_d} a_{j_1 j_2.. j_d}\,f_{j_2}(x_2)\ldots f_{j_d}(x_d)}
{\sum_{j_2\ldots j_d} a_{0 j_2.. j_d}\, f_{j_2}(x_2)\ldots f_{j_d}(x_d)} \ee
as $\int_0^1 f_i(x)dx=\delta_{i0}$. Such sums for pairwise dependencies use only two nonzero $j_i$ indexes (input-output), three for triplewise, and so on. Denominator corresponds to normalization, indeed the fraction becomes 1 for $j_1=0$. Examples for $d=2,3$ are shown in Fig. \ref{2d}, \ref{intr} - generally nominator sums over all indexes with the current indexes of predicted variables. Denominator replaces current variables indexes with zeros for normalization, can be removed if having further (inter-layer) normalization. 

Here is example of  analogous prediction of conditional joint distributions for multiple (2) variables:
$$\rho(y,z|x)=\sum_{j_y j_z} f_{j_y}(y) f_{j_z}(z) 
\frac{\sum_{j_x}a_{j_x j_y j_z}\, f_{j_x}(x)}
{\sum_{j_x}a_{j_x 00}\, f_{j_x}(x)}  $$
Working on expected values would remove $y-z$ mixed moments, making $E[y,z|x]=E[y|x] E[z|x]$ (can be different for density). 

Having such conditional distribution, we can for example calculate expected value e.g. to be propagated by neural networks. For polynomial basis \textbf{expected values} of contributions are: $\int_0^1 x f_0(x)dx =1/2, \int_0^1 x f_1(x)dx =1/\sqrt{12}$, and zero for higher moments, leading to formulas including only $i=0,1$ normalization and the first moment as in Fig. \ref{intr}, e.g.:
\be E[x|y,z]=\int_0^1 x\,\rho(x|y,z)dx=\frac{1}{2} +\frac{1}{\sqrt{12}} 
\frac {\sum_{jk} a_{1jk} f_j(y) f_k(z)}{\sum_{jk} a_{0jk} f_j(y) f_k(z)}  \ee
As further there is rather required CDF normalization which both shifts and rescales, in practice it is sufficient to work on such  nominators, marked in Fig. \ref{intr}, \ref{2d}. 

Restricting it to pairwise dependencies: (single variable of input - single variable of output), similarly to KAN we get summation of trained 1-parameter functions: here polynomials (could be different e.g. B-splines like in KAN) + e.g. approximate fixed CDF for normalization: $E[x|y,z]\propto$
\be {\sum_{jk} a_{1jk} f_j(y) f_k(z)} \xrightarrow[\textrm{KAN-like}]{\textrm{pairwise only}} \sum_j a_{1j0} f_j(y) +a_{10j} f_j(z) \label{KANl}
\ee
\noindent

\begin{figure}[t!]
    \centering
        \includegraphics[width=9cm]{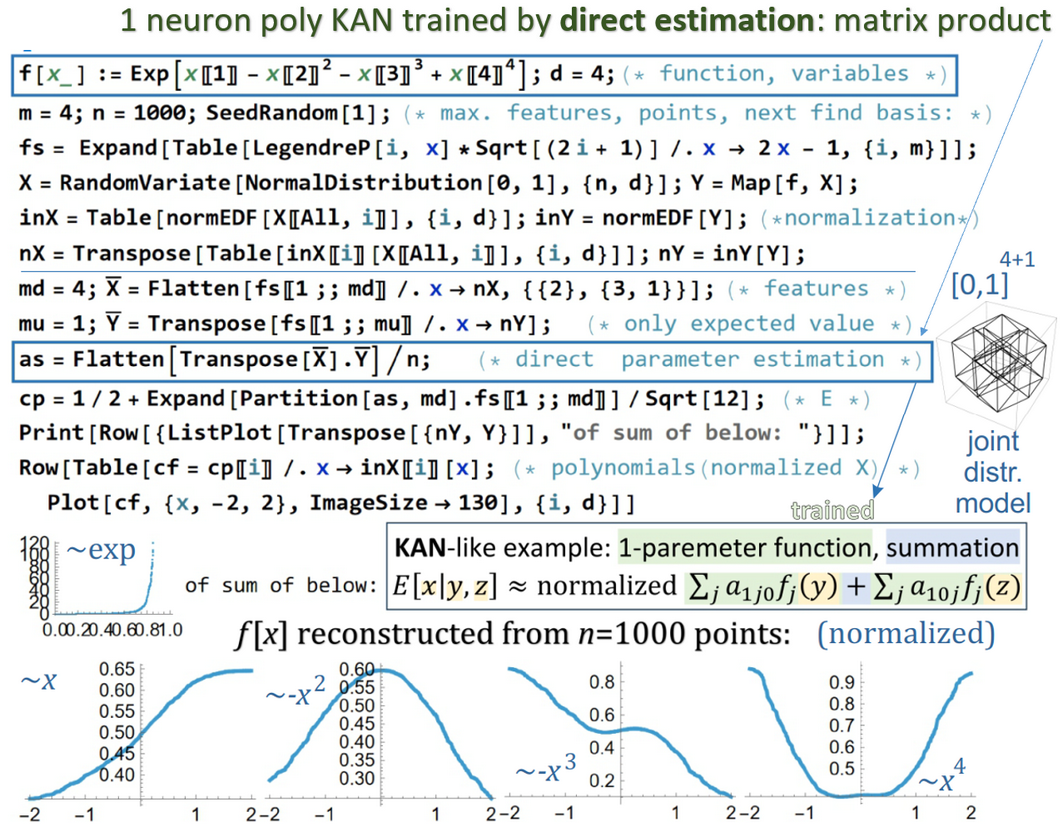}
        \caption{KAN-like example with code from \href{https://community.wolfram.com/groups/-/m/t/3241700}{HCRNN Wolfram notebook}: trying to find $f(x)=\exp(x_1^2-x_2^2-x_3^3+x_4^4)$ from size 1000 random dataset, using single neuron and direct estimation. We can see it has automatically found the four polynomials hidden behind exponent, however, they are deformed as having values normalized to $[0,1]$ range. }
        \label{KANlike}
\end{figure}

\noindent\fbox{%
    \parbox{\columnwidth}{
However, \textbf{in comparison to KAN}, using the proposed HCRNN parametrization we get multiple advantages:
\begin{itemize}
  \item it can \textbf{propagate in any direction} (as BNNs),
  \item propagate values or \textbf{probability distributions} (as BNNs),
  \item interpretation of parameters as \textbf{mixed moments},
  \item consciously add \textbf{triplewise and higher order dependencies},
  \item inexpensive evaluation of modeled \textbf{mutual information},
  \item \textbf{additional training} ways (needed for BNNs), e.g. direct estimation, tensor decomposition, information bottleneck. 
\end{itemize}}}\\

\begin{figure}[t!]
    \centering
        \includegraphics[width=9cm]{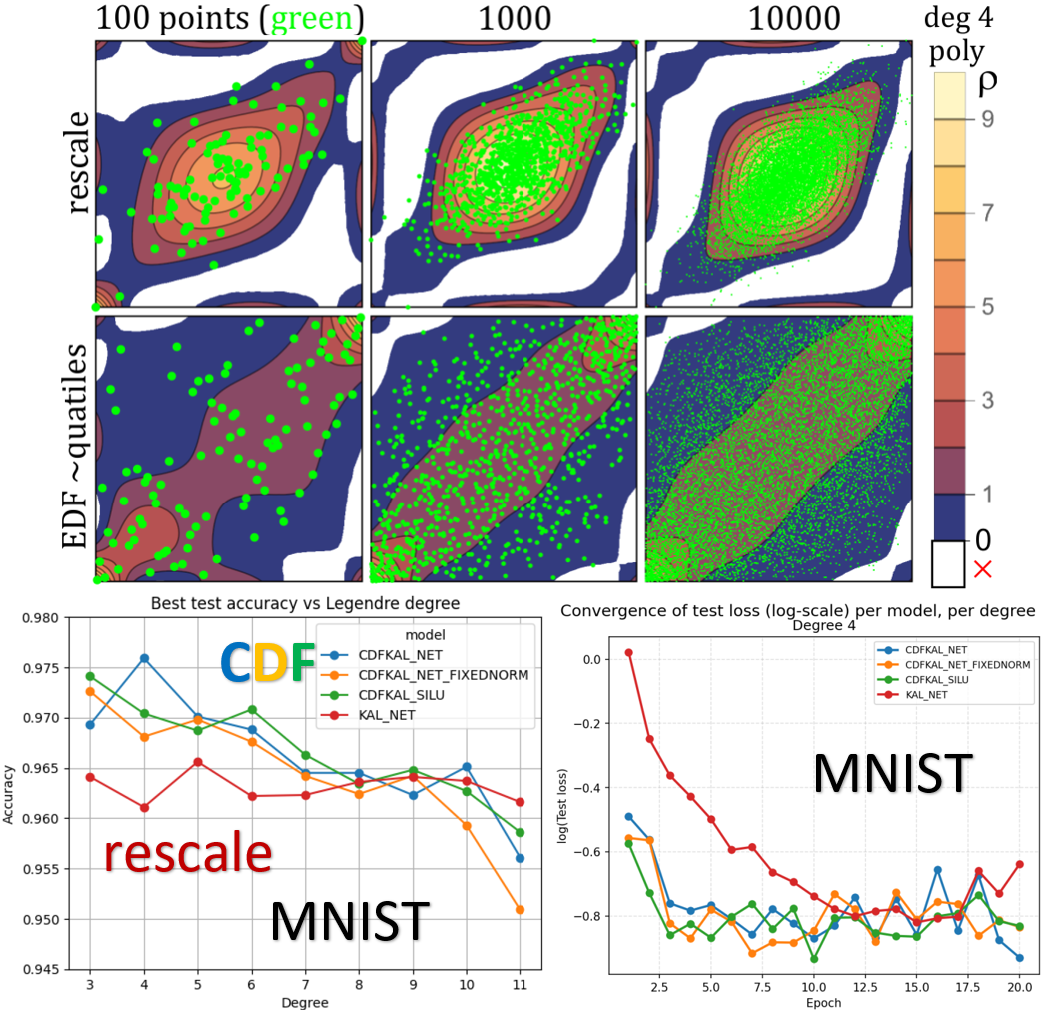}
        \caption{Basic HCR-like improvement of polynomial KAN by just replacing standard rescaling normalization with CDF/EDF, what assuming Gaussian distribution means just adding fixed $\textrm{CDF}_{N(0,1)}$ 
         after standard: mean subtraction and division by standard deviation. At the top we can see that such quantile normalization leads to more uniform distribution - allowing for better representation with low degree polynomials, which are better at generalization. At the bottom there are results from \cite{impKAN} showing such improvements in cross-validation on MNIST for low degree polynomials. For high degree rescale can be better, but their larger number of parameters is more likely to overfit.      
        }
        \label{KANpoly}
\end{figure}

\begin{figure}[t!]
    \centering
        \includegraphics[width=9cm]{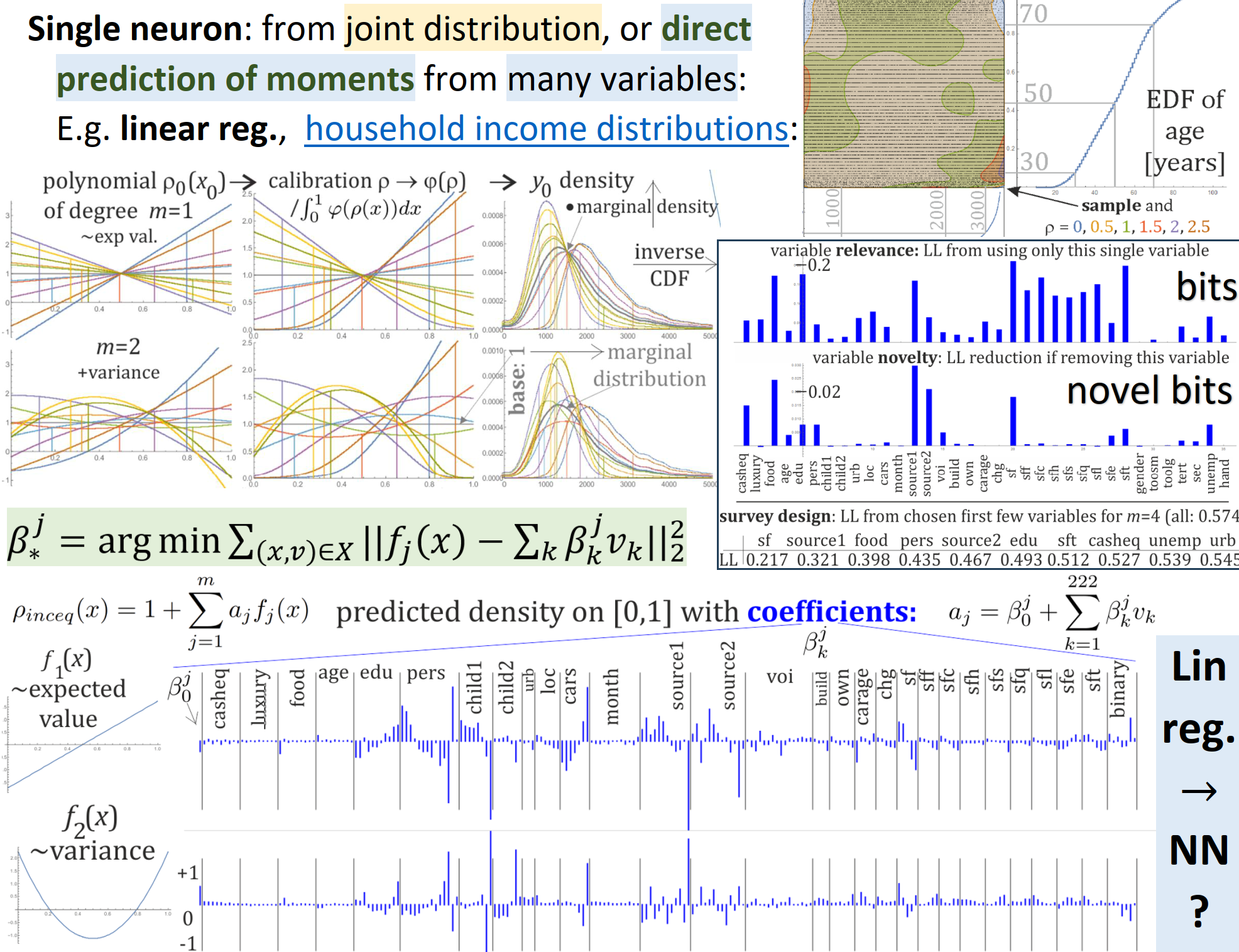}
        \caption{Example of direct prediction of conditional distribution from \cite{cred} - here using separate linear regressions for 1st and 2nd moment $a_j$ with found coefficients visualized at the bottom, from which we get $\rho(x)=\sum_j a_j f_j(x)$ density as polynomial, further requiring calibration to enforce positive density, and finally normalization can be reverted. We get interpretation by contributions to moments, can evaluate variables by relevance: conditional entropy if predicting from single variable, or novelty: entropy drop if removing single variable from their entire set. While least square linear regression was used, it can be naturally extended to neural networks with MSE estimation of conditional moments, e.g. in the last layer of network originally predicting values, or calculating embeddings e.g. in transformer by a few moments representing its probability distribution. }
        \label{direct}
\end{figure}

KAN-like setting as using conditional expected values would include only $(a_{i1})_i$ type coefficients - multiple $i$ powers of inputs to predict 1-st moment of output. Reversing propagation direction, it would predict multiple moments as linearly dependent from value. For more sophisticated dependencies and multidirectional density propagation, we would need to include and train also $(a_{ij})_{ij}$ dependencies between higher moments.

Figure \ref{KANlike} shows single neuron example learning 4 variable function from sample. Figure \ref{KANpoly} shows results from \cite{impKAN} regarding the simplest KAN upgrade by just replacing normalization from rescaling to CDF/EDF. Fig. \ref{direct} shows prediction of conditional distributions from \cite{cred} by separate predictions of multiple conditional moments - can be uses as the last layer of e.g. KAN, or different neural network e.g. for embedding - as a few conditional moments representing its probability distributions.

\subsection{Propagation of probability densities as vectors of moments} \label{densprop}
Let us start with a simple example: that we would like to calculate conditional probability density like previously: 
\be \rho(x|y)=\sum_{i} f_i(x) \frac{\sum_j a_{ij} f_j(y)}{\sum_j a_{0j} f_j(y)}\ee
but now for $y$ being from $\rho(y)=\sum_k b_k\,f_k(y)$ probability density. So the \textbf{propagated probability density} of $x$ should be $\int_0^1 \rho(x|y) \rho(y) dy$. Approximating with constant denominator, using $\int_0^1 f_j(y) f_k(y)dy=\delta_{jk}$ and finally normalizing, we get: 
\be \rho(x)\leftarrow \int_0^1 \rho(x|y) \rho(y) dy \approx  \sum_{i} f_i(x) 
\frac{\sum_j a_{ij} b_{j}}{\sum_j a_{0j} b_{j}}\ee

Such \textbf{constant denominator approximation} allows to propagate (in any direction) through HCR neurons not only values, but also entire probability distributions - by just replacing $f_j(y)$ for concrete value of $y$, with $b_j$ describing its probability distribution. It is natural to generalize, e.g. for   $\rho(x|y,z)$ we could replace $f_j(y) f_k(z)$ with $b_{jk}$ when $\rho(y,z)=\sum_{jk} b_{jk}\, f_j(y) f_k(z)$:
$$ \sum_i f_i(x) \frac{\sum_j a_{ijk}\ f_j(y) f_k(z)}{\sum_j a_{0jk}\  f_j(y) f_k(z)}\quad\stackrel[\textrm{density}]{\textrm{value}}{\leftrightarrows}
\quad\sum_i f_i(x) \frac{\sum_j a_{ijk}\ b_{jk}}{\sum_j a_{0jk}\  b_{jk}}$$

\subsection{Tensor decomposition and linearization for $a_\mathbf{j}\sim f_\mathbf{j}(\mathbf{x})$}
Analogously for intermediate layers like in bottom of Fig. \ref{intr}: 
$$A^k_{i_1i_2i_3i_4}\approx \sum_{j_1,j_2} a^{j_1}_{i_1i_2}\,b^{j_2}_{i_3i_4}c^{k}_{j_1j_2}\qquad\textrm{as}\qquad
\begin{array}{c} c^k_{j_1,j_2} \\\Sigma_{j_1} \wedge\Sigma_{j_2}\\ a_{i_1i_2}^{j_1}\quad  b_{i_3i_4}^{j_2} \\   \end{array}  $$
integrating over the intermediate variables, approximating with constant denominator and normalizing at the end, thanks to basis orthogonality we get Kronecker deltas enforcing equality of intermediate indexes, leading to condition for approximation of higher order tensors with lower order ones, which is generally studied by \textbf{tensor decomposition} field~\cite{tensor} generalizing SVD (singular value decomposition) - what might lead to better training approaches. We can also extend it with nonlinearities to neural networks - optimizing final evaluation, e.g. MSE conditional moments or log-likelihood, at worst it will lead to linear.

While neural networks require nonlinearity, such tensor approach allows to \textbf{linearize} its intrinsic behavior: calculate nonlinearities in some basis e.g. polynomials only for the outer inputs/outputs $(f_\mathbf{j}(\mathbf{x}):\mathbf{j}\in B, \mathbf{x}\in \bar{X})$, including multivariate dependencies. Then treat the entire neural network as a linear transformation of such features (normalization only at both ends - no need for inter-layer), e.g. just changing indexes (like transposition) to modify propagation direction. However, it contains this constant denominator approximation, and it would become a tensor of exponentially increasing size if including all dependencies - it should be combined with reductions like tensor decomposition (into neural network), information bottleneck - working on linear approximations to reduce dimension.

\subsection{Basis optimization and selection}
Another direction is application of the found $a_\mathbf{j}$ coefficients, for example to optimize the arbitrarily chosen $\{f_i\}$ basis to be able to reduce the number of considered coefficients, also to reduce overfitting issues, e.g. discussed in \cite{hcr5,basis}. For this purpose we can for example treat the current coefficients as a rectangular matrix $M_{j_1,(j_2..j_d)}:=a_\mathbf{j}$ - with blocked all but one  indexes for all the considered coefficients in the basis. Now we can use SVD(singular value decomposition)/CCA: find orthonormal eigenbasis of $M M^T=\sum_i \sigma_i \mathbf{u}_i \mathbf{u}_i^T$ and use $g_i =\sum_j u_{ij} f_j$ as the new basis for one or a few dominant eigenvectors. Similarly we can do for the remaining variables, getting separate or common \textbf{optimized bases} for them. \\

A more difficult question is \textbf{basis selection} - which $\mathbf{j}\in B$ indexes to use in considered linear combinations for each neuron. Extending all to $m$-th moment/degree for $d$ variables, we would need $(m+1)^d=\sum_{k=0}^d {d\choose k} m^k$ coefficients: 1 for normalization, $dm$ for marginal distributions, $d(d-1)m^2 /2$ for pairwise, and so on. With proper normalization the coefficients for marginal distributions should be close to 0 - can be neglected. To reduce number of coefficients we can e.g. restrict up to pairwise dependencies ($\approx$ KAN), and/or restrict summed indexes (instead of individual): $\sum_i j_i\leq m$ (polynomial degree). Generally we can calculate more coefficients and discard those close to zero in training, maybe use $l^1$ normalization to find such sparse basis. Using optimized bases as above should allow to reduce $|B|$ size.

\subsection{Some HCRNN training approaches}
A single HCR neuron models multidimensional joint distribution, what is already quite powerful. However, for neural networks the main difficulty is training of the intermediate layers. HCRNN is very flexible also here, below are some approaches:
\begin{itemize}
  \item Treat HCRNN as just a parametrization and use standard backpropagation as for MLP, KAN, e.g. with some distance for values, or log-likelihood evaluation for density propagation. It can be mixed with other techniques, e.g. static parameter estimation/update from recent values, online basis optimization and selection, or just optimizing initial parameters to improve further main  optimization.
  \item Maybe find initial intermediate values by dimensionality reduction like PCA of $\{f_\mathbf{j}(x):\mathbf{j}\in B\}$ vectors of features as (nonlinear) products of functions of inputs - further extended into information bottleneck approach.
  \item Maybe use propagation in both directions and combine with direct coefficient estimation/update (\ref{estim}).
  \item Maybe use some tensor decomposition techniques - start with estimation of dependencies for a larger set of variables, and use algebraic methods to try to approximate it with multiple lower order tensors.
  \item \textbf{information bottleneck approach} local and looking the most promising we discuss further - optimizing joint distributions between layers, hence perfect for neurons with local joint distributions models we focus on.
\end{itemize}

Coefficients of such trained HCRNN remain mixed moments - providing dependency interpretation between input/output, and hidden intermediate variables, allowing for multidirectional propagation of values or distributions like in Fig. \ref{neuron}, and its parameters can be further continuously updated e.g. using (\ref{update}).

\section{Information bottleneck based training}
Let us consider $l$ hidden layer neural network intended to predict $Y$ from $X$ with $\{\theta^i\}_{i=0..l}$ sets of parameters:
\be X\xrightarrow{\theta^0} T^1\xrightarrow{\theta^1} \ldots \xrightarrow{\theta^{l-1}} T^l \xrightarrow{\theta^l} \hat{Y} \quad
\textrm{predicting }Y \ee
While the standard training approach is focused on optimization of neuron parameters $\theta$ used to process the data, alternatively we could try to \textbf{directly optimize content of $T^i$ hidden/intermediate layers}, in practice: dataset processed through some of layers (here could be in both directions) - like in image recognition: first layers extract low level features like edges, then some intermediate features, and finally e.g. faces.

\textbf{Information bottleneck approach}~(\cite{information,information2}) suggests how to directly optimize the content of intermediate layers. It uses \textbf{information theory} - offering nearly objective evaluation, being invariant to variable permutations and bijections: \textbf{mutual information} $I(X;Y)=H(X)+H(Y)-H(X,Y)$, being the number of bits (or nits) $X$ on average brings about $Y$, or $Y$ about $X$.

Optimizing intermediate layer of $X\to T\to Y$, obviously it should maximize information about the predicted $Y$: $\max_T I(T;Y)$. However, focusing only on \textbf{prediction} would rather lead to overfitting. To prevent it, we can also perform \textbf{compression}: minimize $\min_T I(X;T)$ - removing unnecessary information/noise from the input. Finally, information bottleneck approach, for some $\beta>0$, assumes optimization of both:
\be \inf_T \left(I(X;T) - \beta I(T;Y)\right) \label{est}\ee 
\subsection{Information theory view on HCR}
Mutual information has excellent properties: independence to variable permutation and bijections, says how many bits one variable gives on average about the second. However, it is relatively difficult to calculate, requires a joint distribution model - HCR does as $\rho(\mathbf{x})=\sum_{\mathbf{j}\in B} a_{\mathbf{j}}\, f_{\mathbf{j}}(\mathbf{x})$. Using natural logarithm (information in nits) with 1st order approximation $\ln(1+a)\approx a$, and orthogonality of basis, we get simple practical approximation of \textbf{entropy}, formally all \textbf{differential of normalized variables}:
\be H(X)=-\int_{[0,1]^d}\rho(\mathbf{x})\ln(\rho(\mathbf{x}))d\mathbf{x} \approx -\sum_{\mathbf{j}\in B^+} (a_\mathbf{j})^2   \ee 
Analogously for \textbf{cross entropy} with $\rho_b(\mathbf{x})=\sum_{\mathbf{j}\in B} b_{\mathbf{j}}\, f_{\mathbf{j}}(\mathbf{x})$: 
$$-\int_{[0,1]^d}\rho(\mathbf{x})\ln(\rho_b (\mathbf{x}))d\mathbf{x}\approx  -\sum_{\mathbf{j}\in B^+} a_\mathbf{j} b_\mathbf{j} $$
For joint distribution of $(X,Y)$ variables (can be multivariate), denoting $B_{X}, B_{Y}$ as bases used for individual variables, such approximate formula for \textbf{joint entropy} becomes:
\be H(X,Y) =-\sum_{(\mathbf{j}_x,\mathbf{j}_y)\in B_X \times B_Y \backslash\{\mathbf{0}\}} \left(a_{(\mathbf{j}_x,\mathbf{j}_y)}\right)^2 \ee
For $H(X|Y)$ \textbf{conditional entropy} we subtract $H(Y)$, corresponding to $\{\mathbf{0}\}\times B_Y^+$ summation indexes, leaving $B_X^+\times B_Y$:
$$ H(X|Y) =H(X,Y)-H(Y) \approx - \sum_{\mathbf{j}_x\in B_X^+} \sum_{\mathbf{j}_y \in B_Y} \left(a_{(\mathbf{j}_x,\mathbf{j}_y)}\right)^2 $$
Finally for \textbf{mutual information} we remove first row and column:
\be I(X;Y)=H(X)+H(Y)-H(X,Y) \approx \sum_{\mathbf{j}_x\in B_{X}^+}\  \sum_{\mathbf{j}_y\in B_{X}^+} \left(a_{(\mathbf{j}_x,\mathbf{j}_y)}\right)^2 \label{mif} \ee 
hence, as in Fig. \ref{HCR}, this $\ln(1+a)\approx a$ approximation allows to evaluate mutual information by just summing squared nontrivial coefficients: mixed moments between the two variables. 

However, as discussed in \cite{HSICHCR}, square of estimated mean is larger than square of mean by variance divided by sample size $n$, suggesting \textbf{corrected mutual information estimator}:
\be I_c(\hat{X};\hat{Y})=\sum_{\mathbf{j}\in B_X^+}\  \sum_{\mathbf{k}\in B_Y^+} \left(\textrm{mean}(f_\textbf{j} (\textbf{x}) f_\textbf{k}(\textbf{y}))^2-\frac{\textrm{var}(f_\textbf{j} (\textbf{x}) f_\textbf{k}(\textbf{y}))}{n}\right) \label{mifc} \ee

\begin{figure}[b!]
    \centering
        \includegraphics[width=9cm]{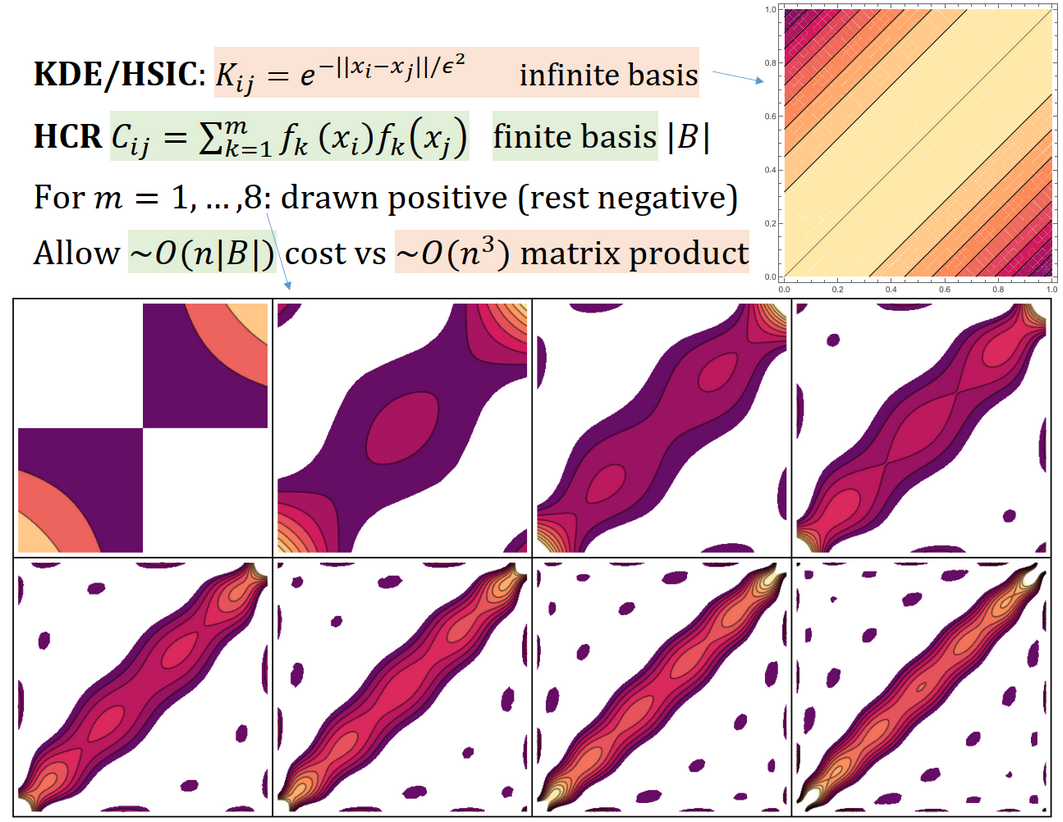}
        \caption{While HSIC (Hilbert-Schmidt Information Criterion) e.g. in \textbf{information bottleneck training}~\cite{hsic2} usually uses Gaussian kernel, proposed HCR approach can be also viewed as kernel method - we can see better localized in diagonal of $[0,1]^2$ for normalized variables. While HSIC has infinite number of Gaussian centers, in HCR we use finite basis - allowing to avoid multiplication of $n\times n$ matrices for size $n$ data sample of  $O(n^3)$ or currently the lowest $O(n^{2.37})$ time complexity, reducing it to linear $O(n|B|)$ for $B$ chosen basis.
        }
        \label{kernels}
\end{figure}

\begin{figure}[b!]
    \centering
        \includegraphics[width=9cm]{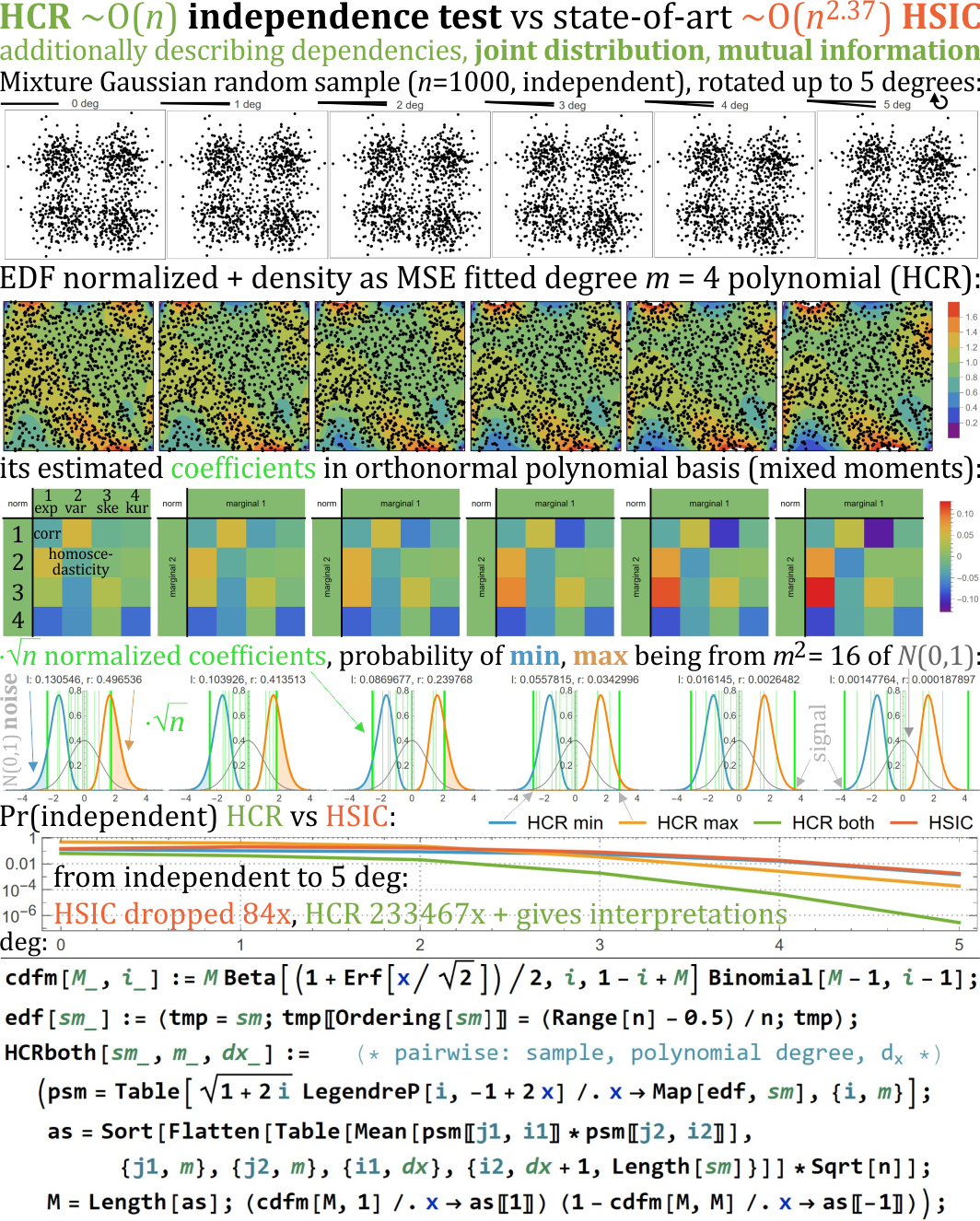}
        \caption{\textbf{Independence test HSIC vs HCR comparison} from \cite{HSICHCR} - we independently generate 2 data samples from bimodal distribution and introduce dependence by rotating it 0, 1, 2, 3, 4, 5 degrees (top). In HCR we model their joint distribution as polynomial - there are shown such density models and their $|B|=4\times 4=16$ coefficients for $m=4$. To distinguish signal from noise, these moments can be normalized to $N(0,1)$ assuming \textbf{independence hypothesis}, hence we can use normality test for such 16 coefficients. The most promising turned out testing the lowest and the highest (showing main dependencies), comparing them with distributions of these extreme values for 16 i.i.d. $N(0,1)$ variables. Here it turned out \textbf{much more sensitive to dependencies than HSIC, also much less computationally expensive, and additionally providing description of the found dependencies}.}
        \label{hsichcr}
\end{figure}

\subsection{Kernel formulation like HSIC independence test}
Let us now go toward HSIC~\cite{HSIC0} kernel formulations, for simplicity without the above correction. Denoting $\{\mathbf{x}^i\}_{i=1..n}, \{\mathbf{y}^i\}_{i=1..n}$ as the current batch from size $n$ data sample, denote 
\be\bar{X}=\frac{1}{\sqrt{n}} (f_\mathbf{j}(\mathbf{x}^i))_{i=1..n,j\in B^+_X}\qquad \bar{Y}=\frac{1}{\sqrt{n}}(f_\mathbf{j}(\mathbf{y}^i))_{i=1..n,j\in B^+_Y}\ee
as $n\times |B^+_X|, n\times |B^+_Y|$ matrices containing features of  vectors from the batch - \textbf{values in the chosen basis} (e.g. with only single nonzero indexes for KAN-like).

Having above $\bar{X}, \bar{Y}$ matrices of features, we can directly MSE estimate (\ref{estim}) parameters for all  $B_{X}^+ \times B_{Y}^+$ pairs forming $\theta_{XY}$ matrix. Using $\|A\|_F^2=\sum_{ij} (A_{ij})^2 =\textrm{Tr}(AA^T)$ and $\textrm{Tr}(AA')=\textrm{Tr}(A'A)$, mutual information (\ref{mif}) becomes:
$$\theta_{XY}:= (a_{(\mathbf{j}_x,\mathbf{j}_y)}: \mathbf{j}_x\in B_X^+, \mathbf{j}_y\in B_{Y}^+) =\bar{X}^T \bar{Y} $$
$$ \qquad I(X;Y)\approx \|\bar{X}^T \bar{Y}\|^2_F= \textrm{Tr}\left(\bar{X}^T \bar{Y} (\bar{X}^T \bar{Y})^T \right) =\textrm{Tr}(C_X C_Y) $$
\be \textrm{for }n\times n \textrm{ \textbf{kernels}:}\quad C_X=\bar{X}\, \bar{X}^T \qquad C_Y=\bar{Y}\, \bar{Y}^T\ee containing scalar products of points in batch as vectors of features. If subtracting the means earlier, they resemble covariance matrices of the used features - this subtraction might be worth including, but generally the means should be close to zero. Also, instead of covariance matrix, this is rather $n\times n$ similarity matrix inside the size $n$  batch using the chosen features. 

Figure \ref{kernels} shows some their values and comparison with Gaussian kernel popular in HSIC, then Fig. \ref{hsichcr} shows their simple benchmark for the original application: independence test, where HCR turned out not only more sensitive to dependencies, and actually allowing to provide their descriptions, but is also much cheaper - allowing to avoid costly multiplication of $n\times n$ kernels by directly using $I(X;Y)\approx \|\bar{X}^T \bar{Y}\|^2_F$ formula of $O(n|B_X^+||B_Y^+|)$ cost of mainly estimation.

\subsection{Gradient descent optimization of information bottleneck}
The previous version of this article contains analytical solution for linear case, but in practice such training rather requires gradient as in  HSIC articles (\cite{hsic1, hsic2}) we will focus on.

Denote $X\xrightarrow{\theta_{XT}} T\xrightarrow{\theta_{TY}} Y$ as the values in 3 consecutive layers: of $n\times n_X$, $n\times n_T$, $n\times n_Y$ dimensions for $n$ size of dataset and $n_X,n_T,n_Y$ numbers of neurons. The optimized $\theta$ coefficients are linear dependencies between moments. For simplicity we assume being fully connected, in other case we can consider $X\to T\to Y$ triples e.g. for individual neurons in $T$, with its connections in $X$ and $Y$.

For nonlinearities we find values in the basis:
$$n\times n_Tm\ \textrm{matrix:}\quad \bar{T}_m=\frac{1}{\sqrt{n}}(f_p(t^i_j))_{i=1..n,\ (j,p)\in (1..n_T)\times (1..m)}$$
and the same for $\bar{X}_m,\bar{Y}_m$. Generally we can use different basis toward left and right: $\bar{T}_I$ for input and $\bar{T}_O$ for output for some natural numbers $I,O \geq 1$, which generally can vary between layers and neurons. It means $O$ moments of deeper/earlier layer are used to predict $I$ moments of further layer, or the opposite if reversing propagation direction. 

For KAN-like propagation of expected values we can choose $I=1$, or higher for propagation of probability distributions, e.g. $I=O$ being 2 to include variances, 3 skewness, 4 kurtosis. While KAN here is unidirectional "multiple moments $\to$ expected value" propagation, coming from HCRNN we can still reverse its direction - directly using "expected value $\to$ multiple moments".

We would like to optimize $\bar{T}$ content of intermediate layer:
\be \bar{X}_O \xrightarrow{\theta_{XT}=\bar{X}_O^T \bar{T}_I}  \left(\bar{T}_I\xleftarrow{f_{1..I}}\bar{T}
\xrightarrow{f_{1..O}}\bar{T}_O\right)
\xrightarrow{\theta_{TY}=\bar{T}_O^T \bar{Y}_I} \bar{Y}_I \ee
where $\theta_{XT}=\bar{X}_O^T \bar{T}_I$ and $\theta_{TY}=\bar{T}_O^T \bar{Y}_I$ are directly MSE estimated parameters.

The chosen basis leads to mutual information approximation:
\be I(X;T) \approx \textrm{Tr}(C_{X_O} C_{T_I})\quad\quad I(T;Y)\approx \textrm{Tr}(C_{T_O} C_{Y_I})\ee
which can be used e.g. for \textbf{information bottleneck formula}:
\be \inf_{(t_j^i):i=1..n,\,j:1..n_T} (\textrm{Tr}(C_{X_O} C_{T_I}) -\beta \textrm{Tr}(C_{T_O} C_{Y_I})) \label{ib2} \ee

Let us find formula for this gradient starting with $C_T, C_{T_m}$, for $f_p$ the considered orthonormal basis, $p=1,\ldots,m$:
$$\frac{\partial (C_T)_{kl}}{\partial t_j^i}=
\frac{\partial (\bar{T}\bar{T}^T)_{kl}}{\partial t_j^i}=
\frac{\partial \sum_a t^k_a t^l_a}{n\, \partial t_j^i}=
\frac{\delta_{ik} t_j^l + t_j^k \delta_{il}}{n} $$
$$\frac{\partial (C_{T_m})_{kl}}{\partial t_j^i}=
\frac{\partial \sum_{ap} f_p(t^k_a) f_p(t^l_a)}{n\,\partial t_j^i}=
\frac{\partial \sum_{p=1}^m f_p(t^k_j) f_p(t^l_j)}{n\,\partial t_j^i}=$$
$$=\frac{\sum_{p=1}^m \delta_{ik} f'_p(t_j^i) f_p(t_j^l) + f_p(t_j^k) \delta_{il} f'_p(t_j^i)}{n}$$

Mutual information e.g. between $X$ and $T$ is approximated as $\textrm{Tr}(C_X C_T)$ with derivative (using symmetry of $C_X =(C_X)^T$):
$$\frac{\partial\textrm{Tr}(C_X C_T)}{\partial t_j^i}=
\frac{\partial\sum_{kl} C_{Xkl} C_{Tkl}}{\partial t_j^i}
=\sum_{kl} C_{Xkl}\frac{\delta_{ik} t_j^l + t_j^k \delta _{il}}{n}=$$
$$=\frac{1}{n}\sum_l C_{Xil} t_j^l +\frac{1}{n}\sum_k C_{Xki} t_j^k =\frac{2}{n}\sum_k C_{Xik} t_j^k=\frac{2}{\sqrt{n}} (C_X \bar{T})_{ij}  $$


$$\left(\frac{\partial I(X;T)}{\partial t_j^i}\approx \right)\quad\frac{\partial\textrm{Tr}\,(C_X C_{T_m})}{\partial t_j^i}=
\sum_{kl} C_{Xkl} \frac{\partial (C_{T_m})_{kl}}{\partial t_j^i}=
$$
$$=\frac{2}{n}\sum_{kp} C_{Xik}\, f_p(t_j^k) f'_p(t_j^i) =
\frac{2}{\sqrt{n}} \sum_{p=1}^{m}(C_X \tilde{T}_p)_{ij}\, f'_p(t^i_j)$$
for $n\times n_T$ matrix  $\tilde{T}_p=\frac{1}{\sqrt{n}}(f_p(x^i_j))_{i=1..n,j=1..n_T}$ containing single power, while $n\times m n_T$ matrix $\bar{T}_m$ contains all $\tilde{T}_{p=1..m}$. Denoting $\tilde{T}'_p=\frac{1}{\sqrt{n}}(f'_p(t^i_j))_{i=1..n,j=1..n_T}$, the above become coordinatewise matrix multiplication.

The final gradient of optimized (\ref{ib2}) information bottleneck function, also extending $C_{X_O}=\bar{X}_O \bar{X}_O^T, C_{Y_I}=\bar{Y}_I \bar{Y}_I^T$ (e.g. to reduce cost by multiplying by $\tilde{T}_p$ first), becomes:
\be\frac{\partial\,( \textrm{Tr}(C_{X_O} C_{T_I}) -\beta \textrm{Tr}(C_{T_O} C_{Y_I}))}{\partial t_j^i}=\ee
$$\frac{2}{\sqrt{n}} \left(\sum_{p=1}^I \left(\bar{X}_O\, \bar{X}_O^T \tilde{T}_p\right)_{ij} f'_p(t^i_j)-\beta \sum_{p=1}^O \left(\bar{Y}_I \,\bar{Y}_I^T\tilde{T}_p\right)_{ij} f'_p(t^i_j) \right) $$
where generally there can be used various $I,O$ basis size for different layers, neurons. Above multiplications could use e.g.  $(\bar{X}_O\, \bar{X}_O^T) \tilde{T}_p$ or $\bar{X}_O\, (\bar{X}_O^T \tilde{T}_p)$ bracketing - while the former uses $n\times n$ kernel, the latter is less expensive over basis - allowed for HCR, but not HSIC.

For practical trening we need to split dataset into batches applied multiple times (epochs) in some sequence, e.g. starting with random initial values of weights/intermediate layers. There is a large freedom to choose training details, where we can apply two philosophies:

\subsubsection{Modification of weights based on gradients}
We can use the formulas for weight estimation from given batch: $\theta_{XT}=\bar{X}_O^T \bar{T}_I$, $\theta_{TY}=\bar{T}_O^T \bar{Y}_I$, for example to update the actually used weights $\bar{\theta}$, e.g. with exponential moving average: $\bar{\theta} += \eta (\theta -\bar{\theta})$. Alternatively we can use the found $g_{ij}=\partial_{t_{ij}}$ gradient, and apply it to modify parameters using $\bar{T}\to \bar{T}-\alpha g$.

\subsubsection{Modifying content of intermediate layers}
We could use this gradient to directly modify contents of intermediate layers corresponding to datapoints. However, as such normalized variables have to stay in $[0,1]$ range, direct use of gradient descent method could easily leave this range. To prevent that, we can for example stretch the variable to $(-\infty,\infty)$ for gradient descent step, e.g. with $s= \textrm{CDF}^{-1}(t), t=\textrm{CDF}(s)$ using CDF of e.g. Gaussian distribution, transforming these hidden variables to this distribution. Derivative over the stretched variable $s$ needs just additional multiplication by its PDF: $\partial L(\textrm{CDF}(s))/\partial s=L'(t)\,  \textrm{CDF}'(s)$.


\section{Calibration also including normalization} \label{calsec}
Modeling (joint) density as a linear combination, while being very convenient, has issue of sometimes getting below zero. In NN approaches we can usually neglect it, e.g. resulting approximations should be automatically compensated during training.

\begin{figure}[t!]
    \centering
        \includegraphics[width=9cm]{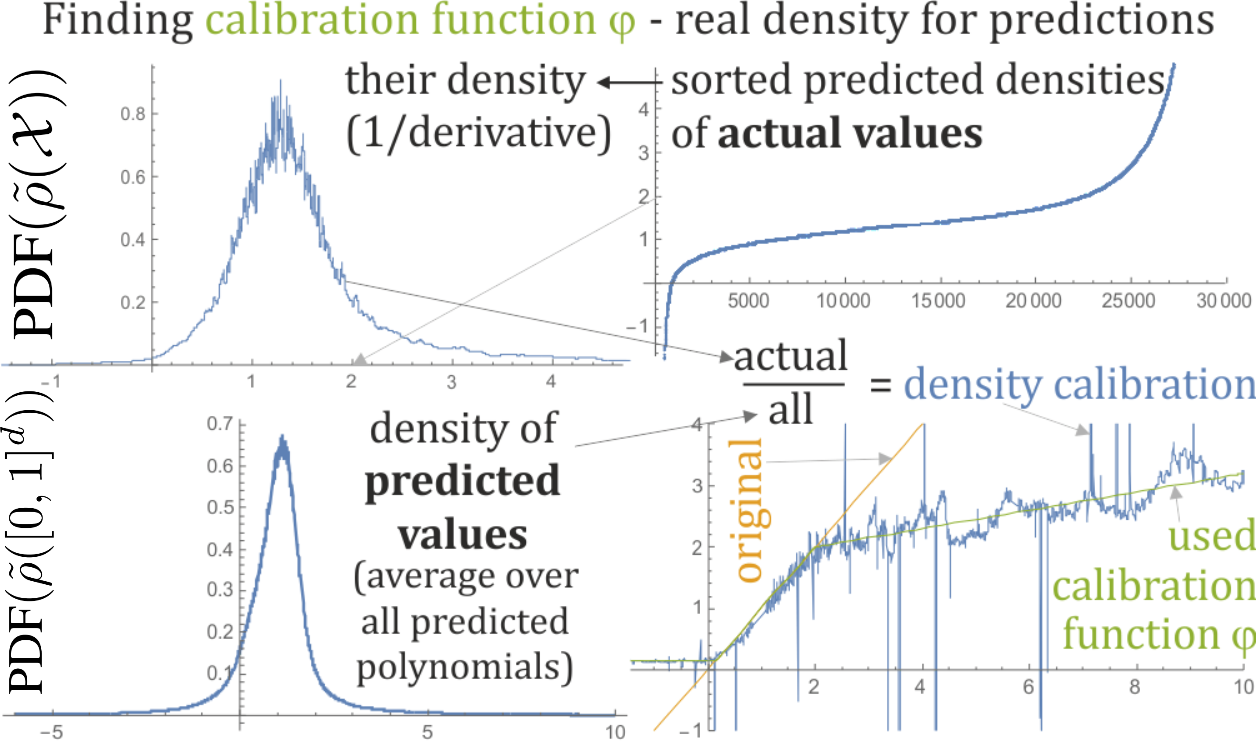}
        \caption{Example from \cite{hcr01} for predicted $d=1$ conditional densities to automatically find calibration functions or its parametrization from data (\ref{calder}): by diving (e.g. KDE) estimated density from dataset $\tilde{\rho}(\mathcal{X})$, by density of all predictions $\tilde{\rho}([0,1]^d)$ e.g. in lattice or on randomly chosen points $\mathcal{P}$. While here for $d=1$ suggests $\varphi_\theta$ parametrization as 3 lines, in higher dimension in Fig. \ref{calex} this approach rather suggests exponential calibration.}
        \label{calib}
\end{figure}

\begin{figure}[t!]
    \centering
        \includegraphics[width=9cm]{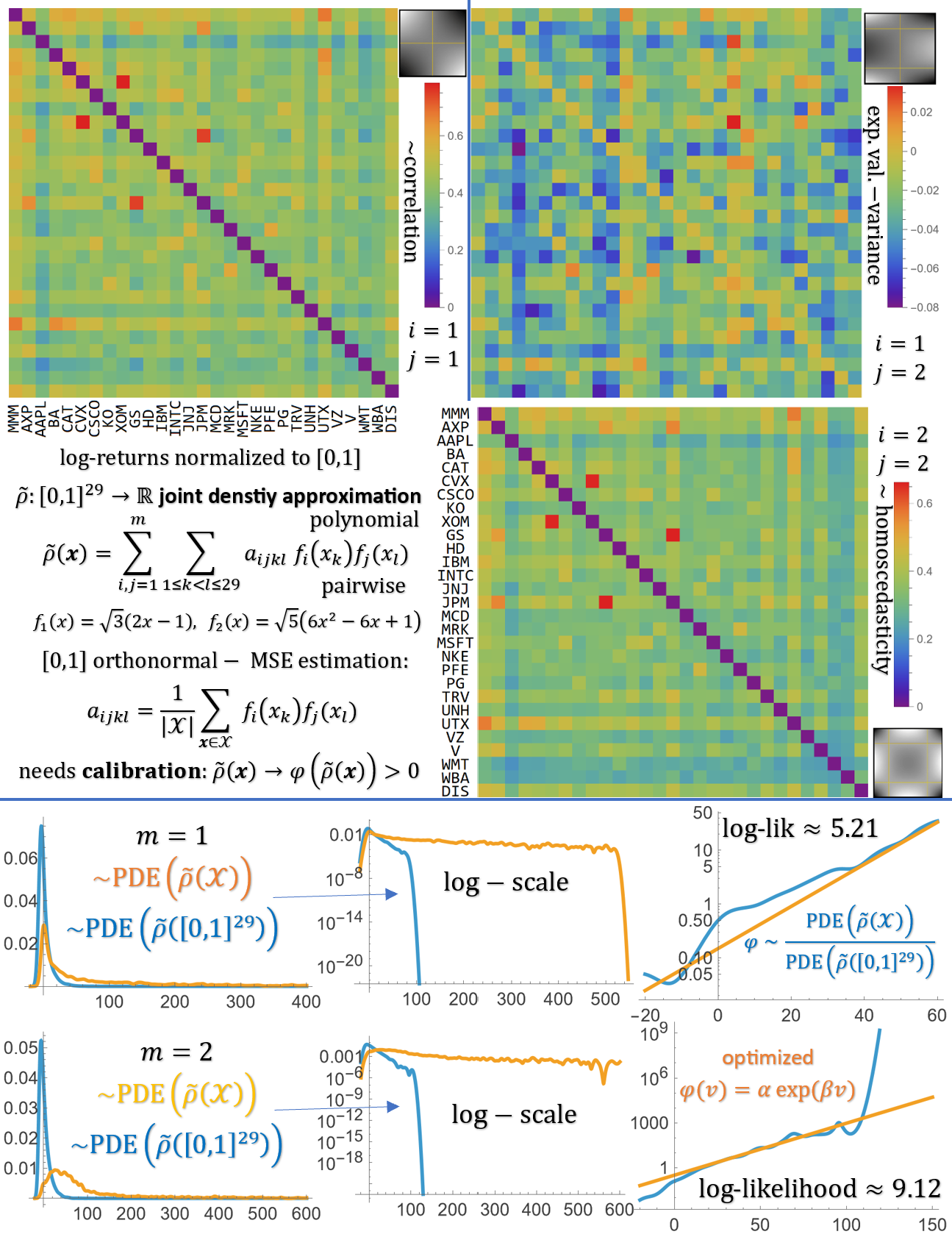}
        \caption{Example of direct HCR modelling of $[0,1]^{29}$ joint density as polynomial of normalized to $[0,1]$ log-returns of 2008-2018 daily prices of stocks of 29 DJIA companies. \textbf{Top}: $29\times 29$ (symmetric) correlation-like moments for $m=1$, and for $m=2$ additional $2\times 29\times 29$ dependencies of variance with first (asymmetric) and second ($\sim$ homoscedasticity) moments.       
        \textbf{Bottom}: while the above moments well describe pairwise dependencies, using them together for high dimensional joint distribution needs reinterpretation of predicted density referred as calibration $\varphi$. Estimating it from data as in Fig. \ref{calib} here suggests  $\varphi(v)=\alpha \exp(\beta v)$ calibration, also further in Fig. \ref{embtab} to agree with softmax. We can first find $\beta$ maximizing log-likelihood in (\ref{opt1}), then $\alpha$ for normalization. }
        \label{calex}
\end{figure}

However, especially for considered the most appropriate (e.g. in MLE): log-likelihood evaluation, its logarithm requires positive density, what can be achieved by calibration: $\varphi:\mathbb{R}\to \mathbb{R}^+$ to reinterpret predicted density $\tilde{\rho}$ into $\varphi(\tilde{\rho})>0$, which can be automatically found from training data as in Fig. \ref{calib}, \ref{calex}.

In practice there are used parameterized e.g. $\varphi(v)=\alpha \max(v,\epsilon)$, soft-plus: $\varphi(v)=\alpha \ln(1 +\beta\exp(\gamma v))$ for some parameters like $\alpha$ for normalization. Here both high-dimensional behavior in Fig. \ref{calex}, and further similarity with transformers suggest to also consider exponential, e.g. $\varphi(v)=\alpha \exp(\beta v)$.

However, then we should ensure normalization: divide $\varphi(\tilde{\rho})$ by integral over $[0,1]^d$ space. This integral in practice is approximated by summation over lattice in $d=1$. In higher dimension we can go iteratively through coordinates: for $\rho(x_1)$, then $\rho(x_2|x_1)$ and so on, or Monte-Carlo approximating the integral by averaging over some large number of randomly chosen points from $\mathcal{P}\subset [0,1]^d$.

Such normalization is impractical e.g. for gradient descent. As in Fig. \ref{direct}, we can separately MSE predict multiple conditional moments, and build conditional distribution from them. We might get better log-likelihood if training these predictors of moments together, however, especially normalization of required calibration would rather make it impractical.

To overcome it, practical approximation is to \textbf{include normalization in calibration function} - like above $\alpha$ parameter locally chosen as constant for $\int_{[0,1]^d} \tilde{\rho}(\mathbf{x})d\mathbf{x}\approx 1$, preferably also maximizing log-likelihood for $\mathcal{X}$ data sample:
\be \arg\max_{\varphi:\mathbb{R}\to\mathbb{R}^+} \left\{\frac{1}{|\mathcal{X}|}\sum_{\mathbf{x}\in\mathcal{X}}\ln(\varphi(\tilde{\rho}(\mathbf{x})))
:  \int_{[0,1]^d} \varphi(\tilde{\rho})(\mathbf{x})d\mathbf{x} =1 \right\}\label{opt0}\ee
In practice we should restrict to some parameterized families $\varphi_\theta$, also can approximate the normalization by averaging over $\mathcal{P}$ lattice or just (Monte-Carlo) set of random points in $[0,1]^d$:
\be \arg\max_{\theta} \left\{\frac{1}{|\mathcal{X}|}\sum_{\mathbf{x}\in \mathcal{X}}\ln(\varphi_\theta(\tilde{\rho}(\mathbf{x}))): \frac{1}{|\mathcal{P}|} \sum_{\mathbf{x}\in \mathcal{P}} \varphi_\theta (\tilde{\rho}(\mathbf{x})))=1   \right\} \label{opt1} \ee

For gradient descent we could regularity update local $\theta$ by (\ref{opt1}) optimization, e.g. replacing its sums by weighted exponential moving averages, and assume $\varphi_\theta(\tilde{\rho}(\mathbf{x}))$ density (including normalization in $\alpha$ scaling parameter) for optimized e.g. log-likelihood evaluation.\\

As mentioned in \cite{hcr01}, we can also directly estimate calibration function by viewing integration of $v=\tilde{\varphi}(\mathbf{x})$ from perspective of entire domain $[0,1]^d$ and real data represented by dataset $\mathcal{X}$:

\be  1=\int_{[0,1]^d}  \rho(\mathbf{x})\, d\mathbf{x} = \int_{[0,1]^d}  \varphi(\tilde{\rho}(\mathbf{x}))\, d\mathbf{x} \overset{v=\tilde{\varphi}(\mathbf{x})}{=} \label{calder}\ee
$$=\int_\mathbb{R} \varphi(v)\, \rho_{[0,1]^d}(v)\, dv=  \int_\mathbb{R} \rho_\mathcal{X}(v)\, dv  \quad \Rightarrow\quad  \varphi(v)\approx \frac{\rho_\mathcal{X}(v)}{\rho_{[0,1]^d}(v)} $$
diving $\rho_\mathcal{X}=\textrm{PDF}(\tilde{\rho}(\mathcal{X}))$ density of data (approximated by dataset $\mathcal{X}$), by density of all predictions $\rho_{[0,1]^d}=\textrm{PDF}(\tilde{\rho}([0,1]^d))$, in practice approximated by lattice or randomly chosen points $\mathcal{P}$. 

These two PDFs can be estimated e.g. with KDE, often leading to noisy behavior like in Fig. \ref{calib}, \ref{calex} - should be rather used to only choose parametrization from $\varphi(v)\approx \rho_\mathcal{X}(v)/\rho_{[0,1]^d}(v)$.

\begin{figure}[t!]
    \centering
        \includegraphics[width=9cm]{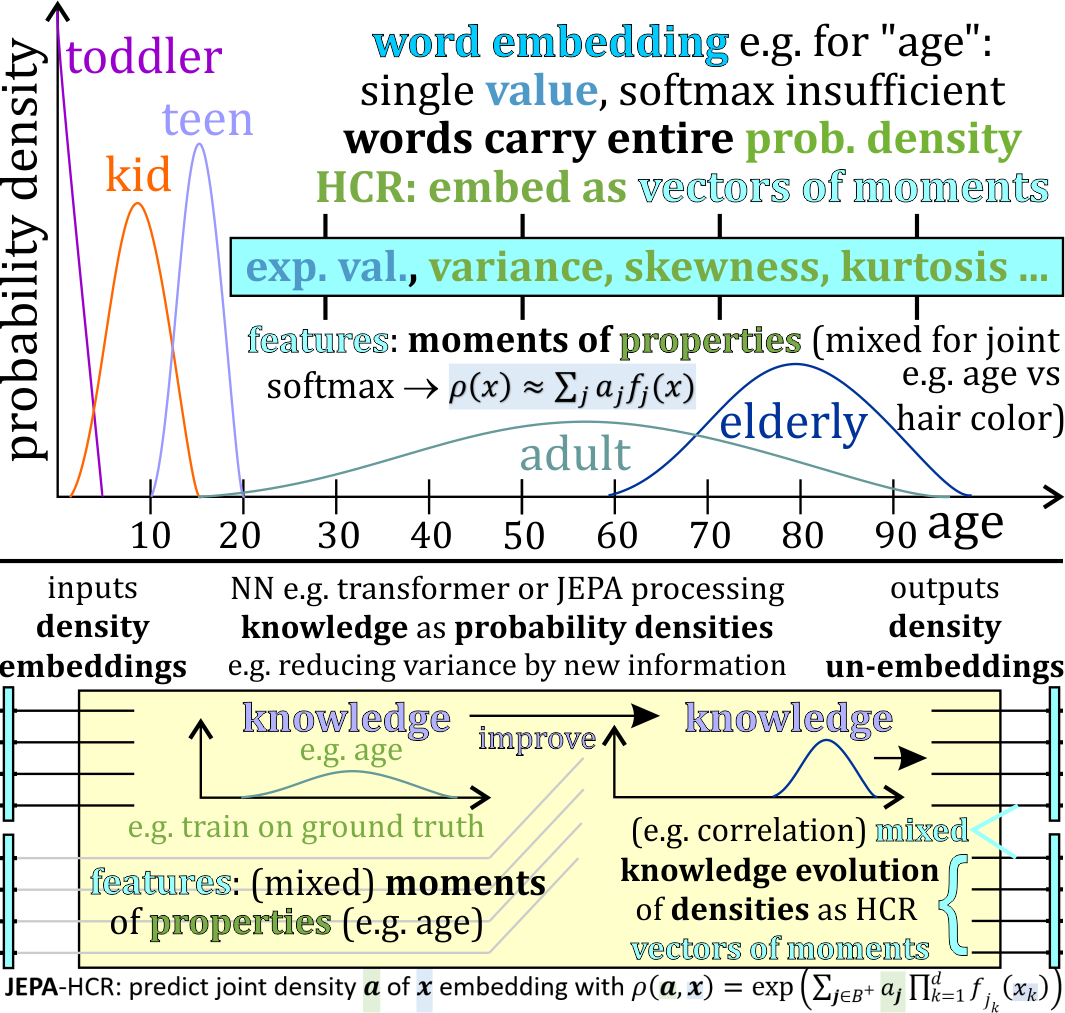}
        \caption{\textbf{Top}: embeddings are the basic tools of modern neural networks like transformers, representing various objects e.g. words as vectors, of parameters hopefully corresponding to real properties, like age. However, e.g. word "adult" represents much larger age variance than "toddler" - single feature like energy in softmax seems insufficient to describe this property, requiring entire probability distributions, e.g. as discussed vectors of moments.
        \textbf{Bottom}: we would like to enforce standard architectures like transformer or JEPA to work on (joint) densities as  internal knowledge, e.g. reducing uncertainty with additional information. Working on vectors of moments, we could group features of embedding (e.g. by 4 here), treating them as moments of some property like age, optionally also including mixed like correlation. For transformers finally there is unembedding, which as in Fig. \ref{embtab} can become popular softmax for features being mixed moments. JEPA uses separate embedding for predicted situations, which could be trained to be real-world parameters, which joint densities can be predicted.}
        \label{embed}
\end{figure}
\section{Extending NN architectures to work on densities} 
Let us now discuss application of presented methodology/philosophy to extend standard NN architectures like JEPA, transformers, Mamba - usually working on values, into working on, predicting probability densities.

\subsection{Density prediction and propagation}
A basic approach is modification of value predictor, into prediction of multiple conditional moments (e.g. extending last NN layer), combining them into prediction of conditional (joint) probability density - like using separate linear regressions in Fig. \ref{direct}, maybe replacing it with neural network, at cost of lower interpretability. Its additional advantage is information theoretic feature evaluation, e.g. estimating numbers of bits available only from given feature - allowing to select such most valuable ones.

With adaptive calibration (\ref{opt1}) we can train conditional moments together (including dependencies) by evaluating the final density with log-likelihood (normalized: in $[0,1]^d$).

While the above can be extension of standard NN propagating values, as discussed in \ref{densprop} we can also directly propagate densities e.g. approximately by just linear transformations there, which could be extended to nonlinearities, their neural networks, allowing them to operate/propage densities e.g. as vectors of moments.

\subsection{Density (un-)embeddings for e.g. transformers and JEPA}
Modern architectures like transformer~\cite{transformer} or JEPA~\cite{JEPA} are often based on embeddings, representing objects e.g. words or images as \textbf{vectors of features}, supposed to describe some their \textbf{real-world properties} (currently nearly not understood).

However, in practice we rather have probability distributions of such properties, which can be improved with further information, allowing to narrow probability distribution of e.g. age while reading some text, or medical parameters with further diagnostics - usually reducing variance. Ideally, we should start with probability distribution of the population, and try to improve its estimation with additional data. As in Fig. \ref{embed}, we could  \textbf{try to enforce features e.g. transformer works on to represent (mixed) moments $\mathbf{a}_\mathbf{j} \approx E_\mathbf{x}[f_\mathbf{j}(\mathbf{x})]$ of real-world properties $\mathbf{x}$}.


As discussed in \ref{densprop}, propagation of densities as vectors of moments $a_\mathbf{j}\sim f_\mathbf{j}(\mathbf{x})$ can be realized already with linear operations, hence could be also extended to nonlinearities in NNs - might be already used by some architectures, with optimized features corresponding to (mixed) moments of some properties.

\begin{figure}[t!]
    \centering
        \includegraphics[width=9cm]{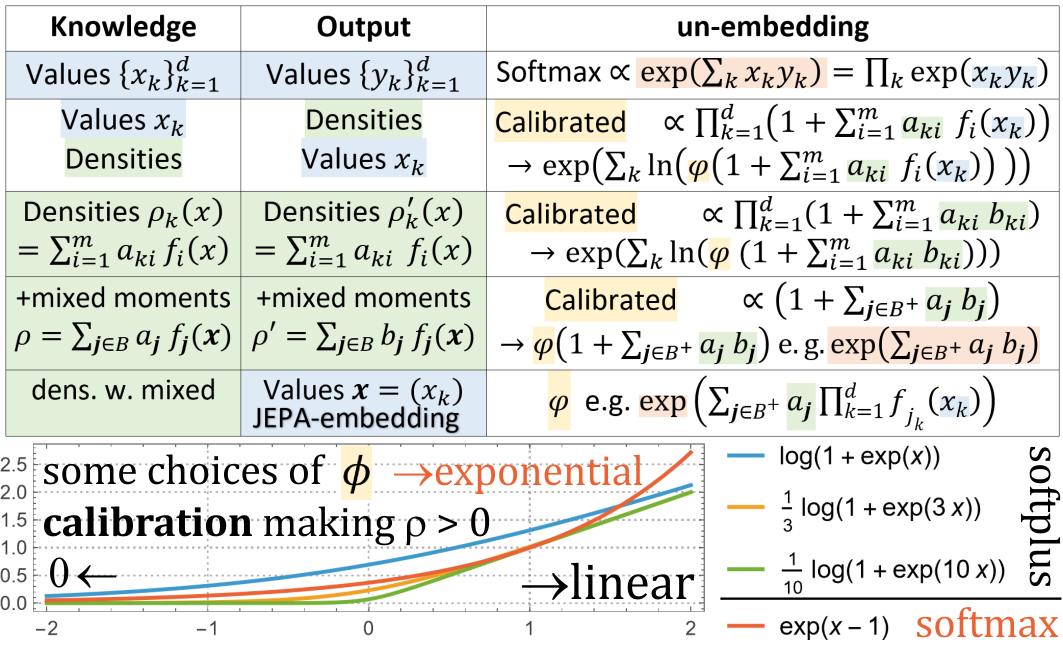}
        \caption{\textbf{Top}: even not having ground truth for properties, we might be able to enforce network to work on their probability densities e.g. through replacement of softmax, un-embedding. There is now mainly used softmax for values (top row). HCR approach allows to replace one or both with density model for each property. There are also dependencies e.g. correlations between properties, suggesting to also include mixed moments between them, getting softmax-like formula, just for mixed moments instead of properties $\mathbf{x}$ - suggesting their interpretation as $f_\mathbf{j}(\mathbf{x})$. \\
        \textbf{Bottom}: density as linear combination might get below zero, what can be prevented by calibration: going through some $\varphi:\mathbb{R} \to \mathbb{R}^+$ function, and performing normalization - which might be omitted as approximation. Choosing it is nontrivial (Section \ref{calsec}), naively should be $\rho(x)\sim x$ for large x like in softplus: $\varphi(x)=\alpha\ln(1+ \beta\exp(\gamma x))$ for some $\alpha,\beta,\gamma>0$. Using $\varphi(x)=\exp(1+x)$ we would get standard softmax for mixed moments, similarity obtained in Fig. \ref{calex} in large dimension, generally it seems worth to test/optimize various $\varphi$.                
        }
        \label{embtab}
\end{figure}

We should also include densities in input/output, e.g. by prediction of probability distribution by a few conditional moments as in Fig. \ref{direct}. To include embeddings, we could try to train them to be (mixed) moments, e.g. based on some ground truth for these properties, e.g. like grokking~\cite{grokking} or from sentiment data.

To try to enforce embeddings as mixed moments, even not knowing of what properties, we could use the final unembedding as in Fig. \ref{embtab} - first working on separate parameters, finally also adding mixed moments, by orthonormality getting trivial:
\be \int_{[0,1]^d} \left(\sum_{\mathbf{j}\in B} a_\mathbf{j} f_\mathbf{j}(\mathbf{x}) \right) \left(\sum_{\mathbf{j}\in B} b_\mathbf{j} f_\mathbf{j}(\mathbf{x}) \right)  d\mathbf{x} = 1+\sum_{\mathbf{j}\in B^+} a_\mathbf{j} b_\mathbf{j} \ee

As linear combination density models can get below 0, we need calibration to make it positive. Using $\varphi(x)=\exp(x-1)=\exp(x)/e$ exponential calibration, which as in Fig. \ref{calex} seems natural in higher dimension, we get the original softmax, just reinterpreting features as mixed moments of parameters.
 
Transformers use embeddings in thousands of features, which might turn out close to of such mixed moments of real-world parameters, leading to better interpretability, and maybe improved architectures. It would be valuable to try to reconstruct the original properties from embeddings, e.g. by help of their ground truth - comparing features with $f_\mathbf{j}(\mathbf{x})$ for $\mathbf{x}$ known properties, maybe using it for initialization or fixing during training. For periodic properties we can switch from polynomial to DCT/DST basis, like in grokking. Finally we should search for automatic extraction of properties, e.g. based on some expected distribution of moments, like higher order ones usually being smaller.

\subsection{JEPA predicting densities of properties of e.g. next frame} 
JEPA (joint embedding predictive architecture)~\cite{JEPA} directly predicts embedding of the next situation (its features), minimizing some distance $\mathcal{D}$ from this prediction. 

For example predicting next video frame, for static objects we are nearly certain their positions, in contrast to dynamical - suggesting to predict (joint) probability densities of such properties like positions. We can extend JEPA to predict next (mixed) moments $\mathbf{a}=(a_{\mathbf{j}})_{\mathbf{j}\in B^+}$ describing marginal distributions ($\#\{i:j_i>0\}=1$), maybe also pair-wise dependencies ($=2$) or higher. It should be trained together with encoder of this next situation into vector of properties $\mathbf{x}\in [0,1]^d$. Then \textbf{density-based distance} between them would be e.g.: minimized $\mathcal{D}(\mathbf{a},\mathbf{x})=1/\rho(\mathbf{a},\mathbf{x})$ or maximized $\ln\rho(\mathbf{a},\mathbf{x})$ for log-likelihood:
\be \rho(\mathbf{a},\mathbf{x})=\varphi\left(\sum_{\mathbf{j}\in B} a_\mathbf{j} f_\mathbf{j}(\mathbf{x})\right)\quad \textrm{for} \quad f_\mathbf{j}(\mathbf{x})=\prod_k f_{j_k}(x_k) \label{dfun}\ee 
with some calibration $\varphi$ e.g. exponent. Such asymmetric $\mathcal{D}$ would allow to additionally \textbf{include nonlinearities} in predictions by $f_j$ polynomials, and hopefully better \textbf{interpretation} - operating on real properties (e.g. positions in video) and their probabilities.

It seems also worth to take (\ref{dfun}) for \textbf{density unembedding} e.g. $\textrm{Pr}(\mathbf{x})= \rho(\mathbf{a},\mathbf{x})/\sum_{\mathbf{x}'} \rho(\mathbf{a},\mathbf{x}')$  also in transformers - to include nonlinearities in prediction, with density interpretation.

JEPA has representation collapse issue, naively preferring same values in embedding. There are various ways to overcome it, here the basic suggested is just batch normalization layer - for each batch spreading all features uniformly over $[0,1]$.

\subsection{Mamba-like extension of adaptive estimation}
Mamba\label{mamb}~\cite{mamba} is recent popular State Space Model, claimed to be more efficient alternative to transformers, as its cost grows linearity with context size, instead of squared for attention. 

It uses trained matrices $A,B,C,D$ for input vectors $x_t\in \mathbb{R}^d$, internal state $h_t$ for time $t$, optimized for agreement of output:
\be \mathbf{h}_{t+1} =A\, \mathbf{h}_t +B\, \mathbf{x}_t\quad \textrm{for output}\qquad \mathbf{y}_t=C\, \mathbf{h}_t +D\, \mathbf{x}_t \ee
It resembles EMA adaptive HCR estimation (\ref{update}), suggesting more educated choice:
\be a_{\mathbf{j},t+1}=\eta_\mathbf{j}\, a_{\mathbf{j},t} + (1-\eta_\mathbf{j})\, f_\mathbf{j}(\mathbf{x}_t)  \ee
We can start with common EMA rate $\eta_\mathbf{j}=\eta\in (0,1)$, then like in Mamba train diagonal of $A$ matrix as individual rates $\eta_\mathbf{j}$. Further we can include also non-diagonal $A$ terms, allowing to exploit dependencies between moments. Finally suggesting discrete and continuous versions: $\mathbf{a}\equiv (a_\mathbf{j}), \mathbf{f}(\mathbf{x})\equiv (f_\mathbf{j} (\mathbf{x}))$ over considered basis of moments $B\ni \mathbf{j}$ (e.g. marginals + pairwise):
\be \mathbf{a}_{t+1}=A\, \mathbf{a}_{t} + B\, \mathbf{f}(\mathbf{x}_t)  \quad\textrm{or}\quad \mathbf{a}'(t)=A\, \mathbf{a}'(t) + B\, \mathbf{f}(\mathbf{x}) \ee
This way we can get evolving density estimation, allowing to replace evaluation as distance between values by  log-likelihood, additionally including e.g. uncertainties in prediction.

We can start with a larger basis $B\ni \mathbf{j}$ and restrict it to the most valuable during training, e.g. based on novelty - lost in bits when removing this moment.

To summarize potential improvements over Mamba:
\begin{itemize}
  \item Replacing minimized distance of predicted values $\mathbf{y}_t$, with prediction of probability densities using log-likelihood evaluation $\sim \sum_\mathbf{j} a_\mathbf{j} f_\mathbf{j}(\mathbf{y}_t)$, optionally with optimized calibration,
  \item Optimized training - e.g. staring with diagonal $A=\eta I$, $B=I-A$, then first training various rates on diagonals $\eta_\mathbf{j}$, and optionally finally freeing non-diagonal, maybe restricted to some family to allow efficient matrix multiplication,  
  \item Optimized nonlinearities in $f_\mathbf{j}(\mathbf{x})=\prod_i f_{j_i}(x_i)$ input/output - instead of guessed standard SiLU,
  \item Better interpretation of coefficients as moments (e.g. marginal + pairwise), prediction as joint distribution.
\end{itemize}

\subsection{Self-supervised learning, e.g. DINO-like~\cite{dino}}
Self-supervised learning is paradigm where the trained model also itself generates supervisory signal, e.g. artificially degrading data, like applying cropping to images already learned to classify - training to still properly classify them, or predict some properties like position, angle, color.

Such applied degradation should reduce certainty, we could include here e.g. with increased variances. For example in DINO~\cite{dino} we directly train student on degraded data, and teacher uses EMA (exponential moving average) of student's weights - in such transfer of weights we could additionally apply e.g. reduction of variances. Or maybe completely remove higher moments in teacher: update its weights from only prediction of expected values in student - treating teacher as value predictor of real-world properties, and student as predictor of their densities.

\section{Conclusions and further work}
\begin{figure}[t!]
    \centering
        \includegraphics[width=9cm]{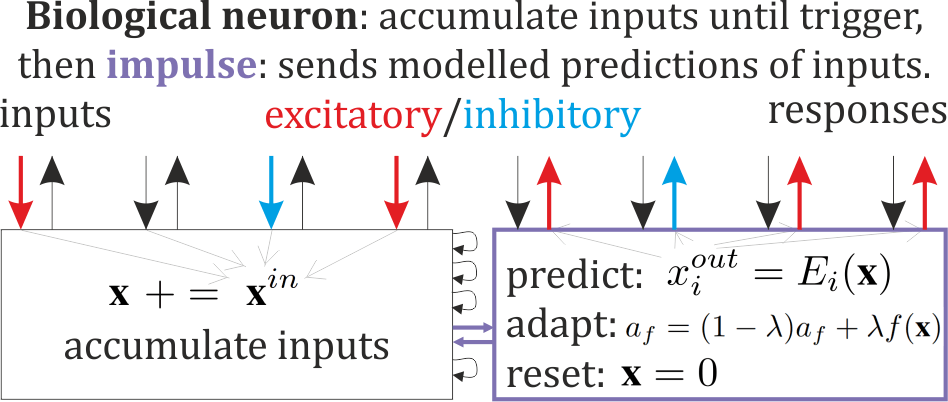}
        \caption{Omnidirectional HCR neuron proposed in \cite{hcr0} - getting any subset $S$ of connections as input, it can update model of joint distribution inside $S$: for $a_\mathbf{j}$ coefficients positive only in this subset $\left(\{i:j_i\geq 1\} \subset S\right)$, and predict/propagate to output as the remaining connections e.g expected values for these inputs, for example accumulated up to some threshold, including sign for excitatory/inhibitory. }
        \label{neuron}
\end{figure}

Neurons with local joint distribution models seem powerful agnostic enhancement for currently popular guessed parametrizations like MLP or KAN, and are practically accessible with HCR, up to omnidirectional neurons like in Fig. \ref{neuron} - allowing to freely choose inference directions, propagate both values and probability distributions, with clear coefficient interpretations as moments, combined into (joint, conditional) distributions when needed.

As in Fig. \ref{table}, \ref{learn}, BNNs are qualitatively superior to current ANNs, also have multidirectional propagation, including probabilities, and need different training than standard backpropagation. Proposed new ANNs allow to catch up with such low level properties, still looking biologically plausible e.g. KAN-like, hopefully allowing to also get closer for high level behavior, maybe recreating mathematics hidden in BNN dynamics.

However, mastering such new neural network architecture will require a lot of work, for example by enhancement of current architectures. Here are some basic further research directions:
\begin{itemize}
\item Search for practical \textbf{applications}, from replacement of standard ANN, for multidirectional inference e.g. in Bayes-like scenarios, as neural networks propagating probability distributions, enhancements of current architectures, up to exploration of similarity/replacement for biological neurons.
\item Practical \textbf{implementation}, \textbf{optimization} especially of training and update, basis optimization and selection techniques, exploration of tensor decomposition approach.
\item Working on probability distributions makes it natural for information theoretic approaches like information bottleneck~\cite{information} optimization of intermediate layers, also hopefully leading to \textbf{better understanding} of information propagation during learning/inference, information held by intermediate layers, embedding features as mixed moments, etc.
\item Adding \textbf{time dependence} like model update, also for similarity with biological neurons, e.g. long term potentiation, connection to various periodic processes/clocks.
\item While the discussed neurons containing joint distribution models seem very powerful and flexible, directly working in high dimensions they have various issues - suggesting to \textbf{directly predict conditional distributions} instead with HCR parametrization (\cite{hcr1,hcr2,hcr4,hcr5}), what might be also worth included in neural network, e.g. as a part of the training process - to be decomposed into single neurons.
\item Modification of softmax e.g. \textbf{embedding} in transformer, JEPA, Mamba approach from single parameter to multiple HCR moments representing density of embedded parameter.
\item Extraction of \textbf{real-world properties} from embeddings interpreted as containing mixed moments, preferably automatic.
\end{itemize}

\bibliographystyle{IEEEtran}
\bibliography{cites}
\end{document}